\documentclass[10pt,twocolumn,letterpaper]{article}

\usepackage[pagenumbers]{wacv}

\usepackage[table]{xcolor}

\definecolor{pearFour}{HTML}{588F27}
\definecolor{pearFith}{HTML}{ECF0F1}
\definecolor{pearDark}{HTML}{2980B9}
\definecolor{pearDarker}{HTML}{1D2DEC}
\definecolor{pearLight}{HTML}{F7E0D5}

\usepackage{url}
\usepackage{nicefrac}
\usepackage{lipsum}
\usepackage{fancyhdr}
\usepackage{appendix}
\usepackage{afterpage}

\usepackage{graphicx}
\usepackage{caption}
\usepackage{subcaption}
\usepackage{wrapfig}
\usepackage{float}
\usepackage{svg}

\usepackage{booktabs}
\usepackage{multirow}
\usepackage{threeparttable}
\usepackage{arydshln}

\usepackage{amsmath}
\usepackage{amssymb}
\usepackage{amsfonts}
\usepackage{bm}
\usepackage{dsfont}

\usepackage{algorithm}
\usepackage{algorithmic}
\usepackage{listings}
\usepackage{verbatim}
\usepackage{comment}

\usepackage{etoolbox}
\makeatletter
\AfterEndEnvironment{algorithm}{\let\@algcomment\relax}
\AtEndEnvironment{algorithm}{\kern2pt\hrule\relax\vskip3pt\@algcomment}
\let\@algcomment\relax
\newcommand\algcomment[1]{\def\@algcomment{\footnotesize#1}}
\renewcommand\fs@ruled{\def\@fs@cfont{\bfseries}\let\@fs@capt\floatc@ruled
  \def\@fs@pre{\hrule height.8pt depth0pt \kern2pt}%
  \def\@fs@post{}%
  \def\@fs@mid{\kern2pt\hrule\kern2pt}%
  \let\@fs@iftopcapt\iftrue}
\makeatother

\usepackage{pifont}

\usepackage{textcomp}
\usepackage{relsize}
\usepackage{xspace}

\usepackage{cancel}

\usepackage{tikz}
\usetikzlibrary{decorations.pathreplacing, positioning}

\definecolor{wacvblue}{rgb}{0.21,0.49,0.74}
\usepackage[pagebackref,breaklinks,colorlinks,allcolors=wacvblue]{hyperref}

\newcommand{\model}{ViTAMINS\xspace}
\newcommand{\vitamin}{\raisebox{-0.2em}{\includegraphics[height=1.2em]{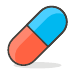}}}

\def\wacvPaperID{770}
\def\confName{WACV}
\def\confYear{2027}

\makeatletter
\apptocmd\@maketitle{{\teaser{}}}{}{}
\makeatother
\newcommand{\teaser}{%
\vspace{-10pt}
\centering
\small
\setlength{\tabcolsep}{2pt}
\begin{tabular}{ccccccccc}
    Image & \texttt{CLS} Att. & Patch Att. & Image & \texttt{CLS} Att. & Patch Att. & Image & \texttt{CLS} Att. & Patch Att. \\
    \includegraphics[width=0.1\linewidth]{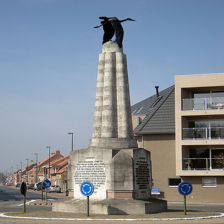} &
    \includegraphics[width=0.1\linewidth]{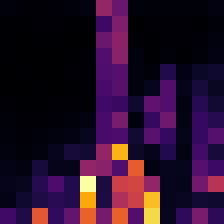} &
    \includegraphics[width=0.1\linewidth]{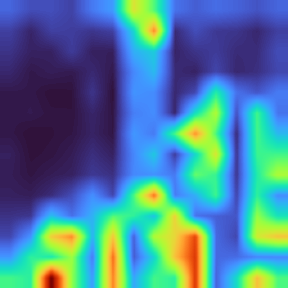} &
    \includegraphics[width=0.1\linewidth]{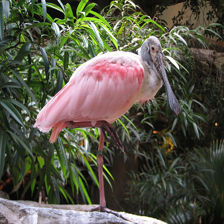} &
    \includegraphics[width=0.1\linewidth]{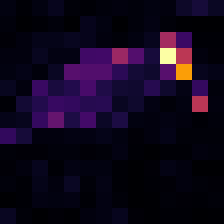} &
    \includegraphics[width=0.1\linewidth]{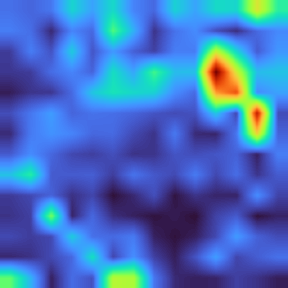} &
    \includegraphics[width=0.1\linewidth]{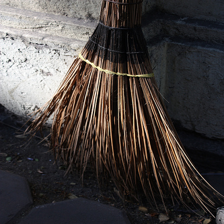} &
    \includegraphics[width=0.1\linewidth]{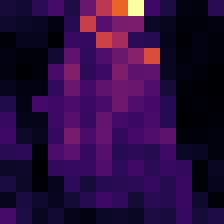} &
    \includegraphics[width=0.1\linewidth]{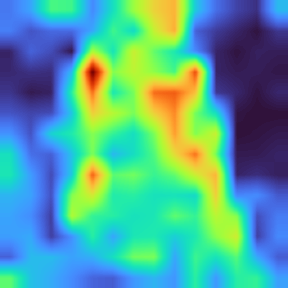} \\
    \includegraphics[width=0.1\linewidth]{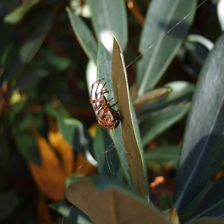} &
    \includegraphics[width=0.1\linewidth]{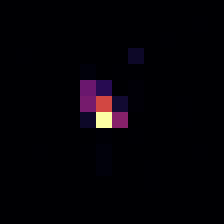} &
    \includegraphics[width=0.1\linewidth]{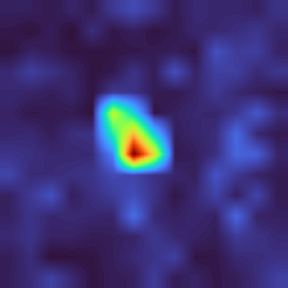} &
    \includegraphics[width=0.1\linewidth]{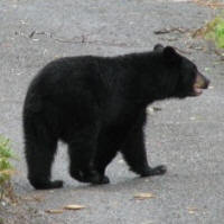} &
    \includegraphics[width=0.1\linewidth]{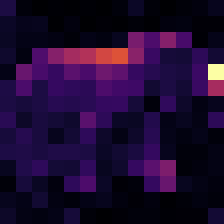} &
    \includegraphics[width=0.1\linewidth]{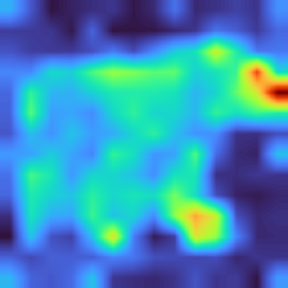} &
    \includegraphics[width=0.1\linewidth]{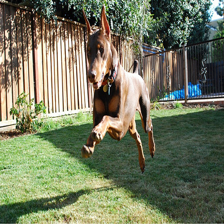} &
    \includegraphics[width=0.1\linewidth]{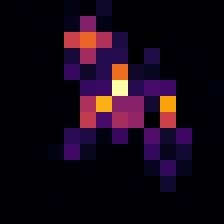} &
    \includegraphics[width=0.1\linewidth]{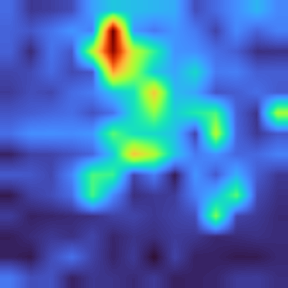} \\
\end{tabular}
\vspace{-6pt}
\captionof{figure}{\textbf{Vision Transformer (ViT-S/16) attention visualization of \model.} For each image set, we show the input image, \texttt{CLS} token attention, and patch attention maps from our method. Our approach with memory and synthetic hard negatives produces focused attention on semantically important regions with clear object boundaries and fine-grained details.}
\label{fig:teaser}
\par\vspace{10pt}
}

\title{
\textit{\model} \vitamin: 
An Empirical Study of Training Self-Supervised Vision Transformers with Synthetic Hard Negatives
}

\author{
Nikos Giakoumoglou \qquad Andreas Floros \qquad Kleanthis-Marios Papadopoulos \qquad Tania Stathaki\\
Imperial College London\\
{\tt\small \{nikos,andreas.floros18,kleanthis-marios.papadopoulos18,tania\}@imperial.ac.uk}\vspace{.5em}\\
{Code: \url{https://github.com/giakoumoglou/vitamins}}\\
}

\begin{document}
\maketitle

\begin{abstract}
We introduce \model, a method that integrates synthetic hard negatives into unsupervised vision transformer pretraining to improve representation quality. Our approach is thoroughly benchmarked on ImageNet and transfer learning, image retrieval, copy detection, and image, video segmentation tasks. Notably, our proposed negatives give rise to \textit{emergent properties}, where learned representations contain explicit information about the semantic content of an image and serve as excellent classifiers (up to +11.3\% over baselines). \model achieves these benefits through simple modifications to existing contrastive frameworks and outperforms competing methods while being more resource efficient, \eg, our ViT-B surpasses V-JEPA with ViT-L. Our findings motivate reconsidering contrastive learning as a simpler yet powerful alternative to dominant generative and self-distillation approaches.
\end{abstract}

\section{Introduction}
\label{sec:introduction}

Transformers \citep{vaswani2017attention} have revolutionized computer vision as powerful alternatives to ConvNets \citep{dosovitskiy2021vit,liu2021swin,touvron2021deit}, adopting an NLP-inspired strategy of pretraining on large data and finetuning on the target dataset \citep{dosovitskiy2021vit,touvron2021deit}. Scaling to billions of parameters on increasingly diverse datasets, they achieve state-of-the-art performance in both supervised and self-supervised paradigms \citep{he2021mae,oquab2023dinov2,goyal2021seer}.

Self-supervised learning has established itself as a powerful approach for visual representation learning, enabling models to extract meaningful patterns from vast amounts of unlabeled data \citep{bommasani2021foundationmodels,lecun2015deep,balestriero2023cookbook,giakoumoglou2024review}. Self-supervised approaches for vision fall into three categories: \textbf{(i)} \textit{pretext task} methods that solve auxiliary tasks such as rotation prediction \citep{gidaris2018rotnet,noroozi2017unsupervised} or jigsaw puzzles \citep{noroozi2017unsupervised}; \textbf{(ii)} \textit{generative} methods that reconstruct or predict masked portions of inputs, such as MAE \citep{he2021mae} inspired by masked language modeling \citep{radford2018improving,devlin2018bert,radford2019language,brown2020language}, and BEiT \citep{bao2022beit} following BERT-like pretraining \citep{devlin2018bert,lan2020albert}; and \textbf{(iii)} \textit{joint embedding architecture} methods that learn representations by comparing different views of data in a shared embedding space \citep{he2020moco,chen2020simclr,grill2020byol,caron2021dino,radford2021clip}. This work focuses on training transformers with joint embedding architectures, unlike prior works using generative methods \citep{he2021mae,bao2022beit,peng2022beitv2}.

\begin{figure*}
    \centering
    {\captionsetup{justification=centering}
    \begin{subfigure}[b]{0.18\linewidth}
        \centering
        \includegraphics[width=\linewidth]{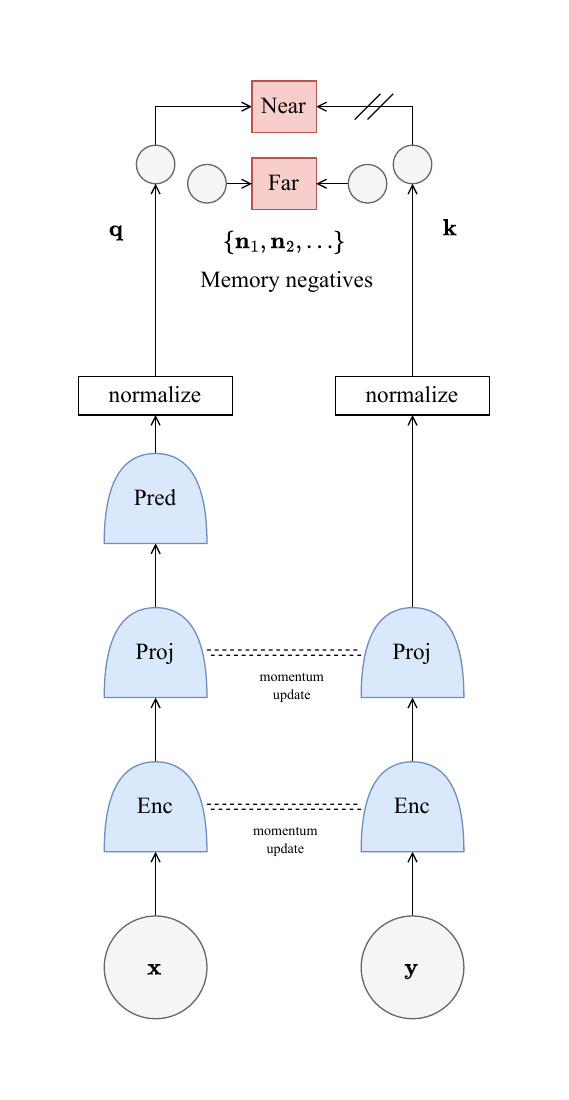}
        \caption{\textit{Contrastive} \\(MoBY)}
        \label{fig:moby}
    \end{subfigure}
    \hfill
    \begin{subfigure}[b]{0.18\linewidth}
        \centering
        \includegraphics[width=\linewidth]{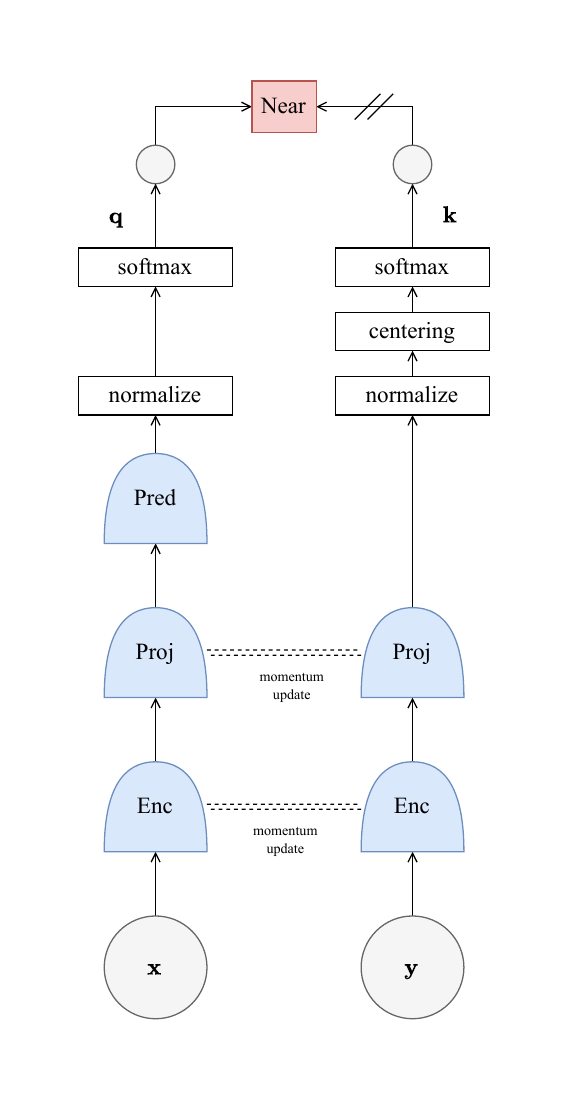}
        \caption{\textit{Self-distillation} \\(DINO)}
        \label{fig:dino}
    \end{subfigure}
    \hfill
    \begin{subfigure}[b]{0.18\linewidth}
        \centering
        \includegraphics[width=\linewidth]{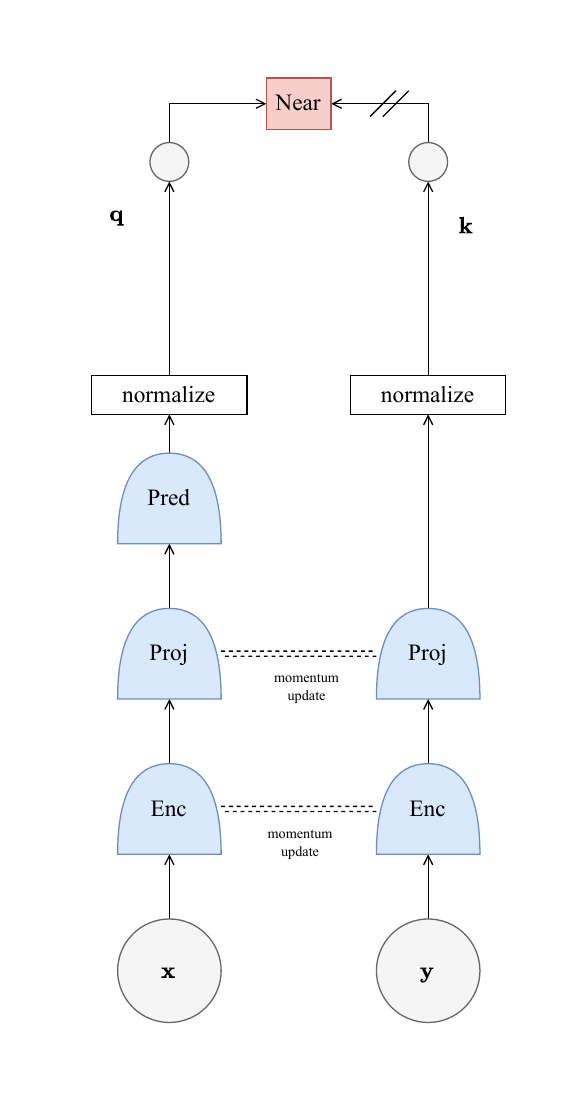}
        \caption{\textit{Self-distillation} \\(BYOL)}
        \label{fig:byol}
    \end{subfigure}
    \hfill
    \begin{subfigure}[b]{0.18\linewidth}
        \centering
        \includegraphics[width=\linewidth]{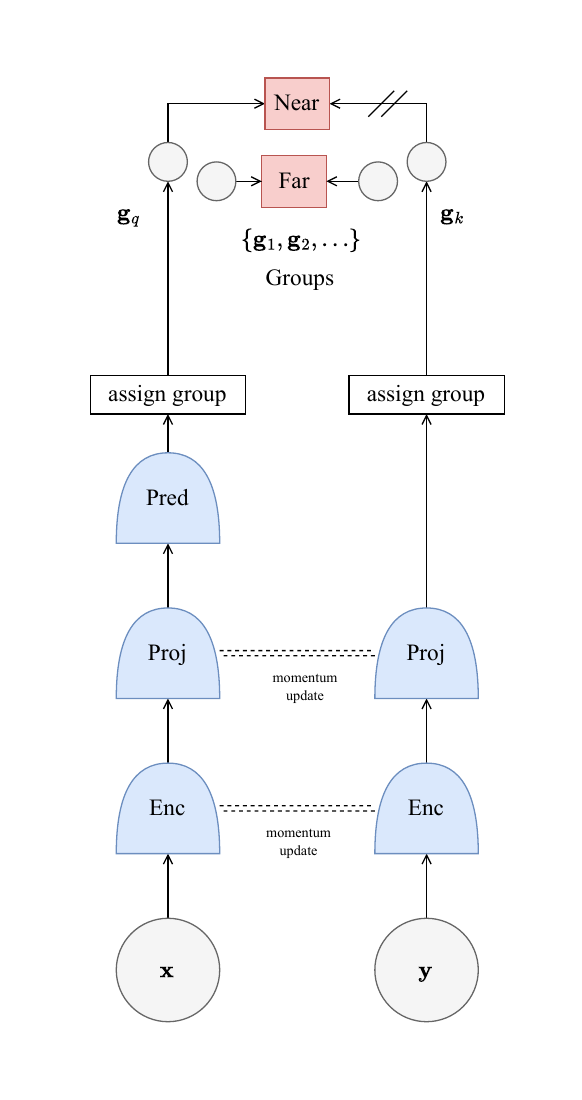}
        \caption{\textit{Clustering} \\(SMoG)}
        \label{fig:smog}
    \end{subfigure}
    \hfill
    \begin{subfigure}[b]{0.18\linewidth}
        \centering
        \includegraphics[width=\linewidth]{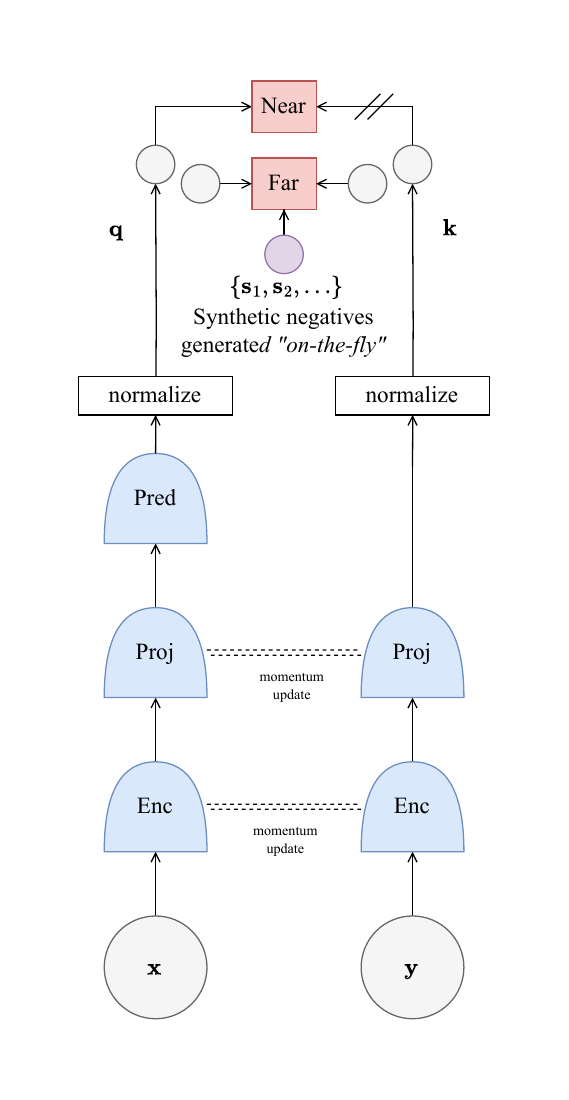}
        \caption{Ours \\(\model)}
        \label{fig:vitamins}
    \end{subfigure}}
    \caption{\textbf{Self-supervised learning categories on vision transformers and this paper's contribution.} From left to right: \subref{fig:moby} \textit{contrastive learning} method \citep{xie2021moby,chen2021mocov3}; \subref{fig:dino}, \subref{fig:byol} \textit{self-distillation} \citep{caron2021dino,grill2020byol}; \subref{fig:smog} \textit{clustering-based} \citep{pang2022smog}; and \subref{fig:vitamins} \model (ours). Our method introduces synthetic hard negatives generated ``\textit{on-the-fly}'' to improve contrastive learning methods for vision transformers. Shaded circles \setlength{\unitlength}{1pt}\begin{picture}(8,8)\put(4,3){\textcolor[HTML]{E6E6E6}{\circle*{8}}}\put(4,3){\circle{8}}\end{picture} represent observed variables, dashed gray lines indicate the momentum update, $\,\mathbin{\!/\mkern-5mu/\!\,}$ indicates a stop-gradient for backpropagation, {\fboxsep 1pt\fcolorbox[HTML]{000000}{FFFFFF}{white boxes}} represent functions, and {\fboxsep 1pt\fcolorbox[HTML]{B85450}{F8CECC}{red boxes}} represent loss functions. \textbf{Abbreviations legend:} Enc: encoder, Proj: projector, Pred: predictor.}
    \label{fig:ssl_categories}
\end{figure*}

The joint embedding methods adapted for vision transformers fall into three categories shown in \Cref{fig:ssl_categories}, each using different ``\textit{tricks}'' to avoid representational collapse: \textbf{(i)} \textit{contrastive learning} methods embed different augmented views of the same image into a joint space, maximizing similarity between same-instance embeddings while minimizing similarity across instances \citep{chen2020simclr,he2020moco,chen2021mocov3,xie2021moby} (\Cref{fig:moby}); \textbf{(ii)} \textit{self-distillation} (\textit{teacher–student}) methods train a student to match a teacher's embeddings without negatives \citep{caron2021dino,grill2020byol,bardes2022vicreg,bardes2024vjepa} (\Cref{fig:dino,fig:byol}); and \textbf{(iii)} \textit{clustering-based} methods employ clustering objectives \citep{caron2019deepcluster,caron2020swav,caron2019deepercluster,pang2022smog} (\Cref{fig:smog}).

Transformers trained with joint embedding methods exhibit emergent properties beyond classification accuracy: their features encode explicit semantic segmentation information that emerges less clearly in supervised transformers or ConvNets \citep{caron2021dino}. Self-distillation methods like DINO \citep{caron2021dino,oquab2023dinov2,simeoni2025dinov3} and iBOT \citep{zhou2022ibot} show strong unsupervised segmentation, with attention aligning to object boundaries without supervision. Such behaviors arise from extensive pretraining and architectural design.

Despite generative methods achieving higher accuracy \citep{he2021mae,bao2022beit}, joint embedding approaches remain competitive and often surpass them in linear probing \citep{oquab2023dinov2,bardes2024vjepa}. Among these, contrastive methods stand out for their simplicity and efficiency \citep{chen2021mocov3,xie2021moby}, explicitly using negatives to define representation boundaries \citep{chen2020simclr,he2020moco}, yet have received less attention recently.

We seek to answer the simple question:
\begin{quote}
\textit{Can simple modifications to negative sampling in contrastive learning unlock stronger representations and emergent properties for vision transformers comparable to or exceeding those of self-distillation methods?}
\end{quote}

In this paper, we address this question by integrating synthetic hard negative generation in transformer-based contrastive learning, a strategy previously demonstrated effective for convolutional networks \citep{giakoumoglou2025synco,kalantidis2020mochi} but \textit{not} investigated for vision transformers. Instead of using complex architectures or training schemes (like multi-crop, centering, sharpening, \etc, see \Cref{sec:related-work}), we adapt established synthetic negative generation approaches to transformer architectures, generating challenging samples ``on-the-fly''.

Through extensive empirical evaluation, we demonstrate that integrating synthetic hard negatives into transformer-based self-supervised learning yields three key improvements over training \textit{without} synthetic hard negatives or \textit{without} negatives: \textbf{(i)} higher top-1 accuracy on ImageNet linear evaluation (\Cref{tab:lineval-imagenet-deit,tab:lineval-imagenet-swin}), achieving 73.1\% and 77.1\% with ViT-S/16 and ViT-B/16, and 75.4\% and 78.0\% with Swin-T and Swin-S, respectively; \textbf{(ii)} improved transfer learning performance across diverse downstream tasks (\Cref{tab:transfer-learning-dense,tab:transfer-learning-linprobe,tab:transfer-learning-ft}); and \textbf{(iii)} strong emergent properties, where self-supervised vision transformer features encode explicit semantic segmentation information (\Cref{tab:retrieval,tab:davis,tab:copydays}), producing precise attention maps that capture object boundaries (\Cref{fig:attention_vit_comparison}) and serve as effective $k$-NN classifiers (\Cref{tab:lineval-imagenet-deit,tab:lineval-imagenet-swin}), achieving 73.3\% top-1 accuracy with ViT-B/16.

\section{Related Work}
\label{sec:related-work}

\subsection{Joint Embedding Architectures}

\textit{Joint embedding architecture} methods map augmented views into a shared embedding space while avoiding representational collapse through distinct mechanisms. \textit{Contrastive learning} methods prevent collapse using large batch sizes \citep{chen2020simclr} or momentum-encoded memory banks \citep{he2020moco,chen2020mocov2,chen2021mocov3,xie2021moby} to provide sufficient negative samples. Alternative approaches formalize collapse avoidance via mutual information \citep{oord2019cpc,hjelm2019dim,tian2020infomin}. \textit{Self-distillation} (\textit{a.k.a.} \textit{teacher-student distillation}) methods surprisingly avoid collapse without negatives. They use asymmetric architectures \citep{chen2020simsiam}, momentum updates \citep{grill2020byol,caron2021dino}, and stop-gradient operations \citep{grill2020byol,caron2021dino,chen2020simsiam}. Alternatively, they explicitly regularize feature covariance so representations do not collapse, \eg, decorrelate features \citep{zbontar2021barlowtwins,bardes2022vicreg}, employ whitening \citep{ermolov2021wmse}, or manifold regularization \citep{yerxa2023mmcr}. Notably, DINO \citep{caron2021dino,oquab2023dinov2,simeoni2025dinov3}, which employs multiple techniques including centering, sharpening, momentum encoder, multi-crop training \citep{caron2020swav}, and extended training, and iBOT \citep{zhou2022ibot}, which integrates masked patch prediction, exhibit strong unsupervised segmentation. Recent predictive architectures extend this by matching embeddings to target distributions; LeJEPA \citep{balestriero2025lejepa} regularizes representations toward an isotropic Gaussian via sketched-based constraints, while Rectified LpJEPA \citep{kuang2026lpjepa} employs rectified distribution matching to enforce sparsity. In contrast, I-JEPA \citep{assran2023ijepa} and V-JEPA \citep{bardes2024vjepa} primarily rely on their masked predictive structure and architectural asymmetry to avoid collapse without explicit variance or covariance constraints. Finally, \textit{clustering-based} approaches align embeddings with prototype assignments obtained via the Sinkhorn-Knopp algorithm \citep{caron2020swav} or via momentum grouping \citep{pang2022smog}. 

\subsection{Contrastive Learning}

\textit{Contrastive learning} methods treat instance discrimination as a pretext task, treating each image as its own class \citep{chen2020simclr,he2020moco}. The core principle involves bringing an anchor and a ``\textit{positive}'' sample closer in the embedding space while pushing the anchor away from ``\textit{negative}'' samples \citep{khosla2021supcon}. Training typically employs InfoNCE loss \citep{oord2019cpc} or its variants \citep{chen2020simclr,dwibedi2021nnclr,tomasev2022relicv2,yeh2022dcl}, maximizing mutual information between positive pairs while minimizing it for negatives. Negative samples are drawn from large batch sizes \citep{chen2020simclr} or memory banks \citep{he2020moco,chen2020mocov2,chen2021mocov3,xie2021moby}. The concept of challenging negative samples has been explored as a way to improve contrastive learning models. These samples, which lie close to the decision boundary, are crucial for refining the model's discriminative abilities \citep{robinson2021contrastive,ali2024overfittingrobustnessquantityquality}. Various strategies leverage hard negatives through mixup-based interpolation between embeddings \citep{kalantidis2020mochi}, debiased contrastive losses with theoretical analysis \citep{robinson2021contrastive}, importance reweighting schemes \citep{yeh2022dcl}, and hardness-aware sampling from memory queues \citep{tomasev2022relicv2}. Systematic synthetic generation through transformation strategies has proven effective for convolutional networks \citep{giakoumoglou2025synco}. Our method adapts synthetic hard negative generation to vision transformers by generating diverse, informative negatives ``on-the-fly'' rather than relying solely on batch size or memory bank capacity.

\section{Methodology}
\label{sec:methodology}

In this section, we introduce our approach, named \model (\underline{\textbf{Vi}}sion \underline{\textbf{T}}r\underline{\textbf{a}}nsfor\underline{\textbf{m}}ers w\underline{\textbf{i}}th sy\underline{\textbf{n}}thetic hard negative\underline{\textbf{s}}). Our method builds upon existing contrastive learning frameworks (see \Cref{fig:moby}) and aims to improve representation quality by generating informative negative samples.

\subsection{\model}
\label{sec:vitamins}

Like other joint embedding methods, \model also operates on the embedding pairs of distorted images. Specifically, given an image $\mathbf{x}$, and two distributions of image augmentation $\mathcal{T}_q, \mathcal{T}_k$, we create two augmented views of the same image using the transformations $t_q \sim \mathcal{T}_q$ and $t_k \sim \mathcal{T}_k$, \textit{i.e}, $\mathbf{x}_q = t_q(x)$ and $\mathbf{x}_k = t_k(x)$.

Then, we use two encoders $f_\theta$ and $f_\xi$, two projectors $g_\theta$ and $g_\xi$, and a predictor $h_\theta$ with parameters $\theta$ and $\xi$ to generate the corresponding embeddings $\mathbf{q}$ and $\mathbf{k}$, where $\mathbf{q} = h_\theta(g_\theta(f_\theta(\mathbf{x}_q)))$ and $\mathbf{k} = g_\xi(f_\xi(\mathbf{x}_k))$, and $\mathbf{q}, \mathbf{k} \in \mathbb{R}^d$ \citep{xie2021moby,grill2020byol}. We denote the \textit{online} branch as $f_\theta$, $g_\theta$, and $h_\theta$, and the \textit{target} branch as $f_\xi$ and $g_\xi$, with parameters $\theta$ and $\xi$, respectively. We assume that the outputs are $\ell_2$-normalized.

We maintain a memory queue $\mathcal{Q} = \left\{\mathbf{n}_1, \ldots, \mathbf{n}_K\right\}$ of features from distinct images serving as $K=4096$ negative samples \citep{he2020moco,chen2020mocov2,chen2021mocov3,xie2021moby}. These $\{\mathbf{n}_i\}_{i=1}^{K}$ are target-branch embeddings from previous steps, requiring memory $\mathcal{O}(K\cdot d)$ with embedding dimension $d$.

Only $\theta$ is updated by backpropagation, while $\xi$ is the exponential moving average $\xi \leftarrow m \cdot \xi + (1 - m) \cdot \theta$, with momentum coefficient $m \in [0,1]$ \citep{grill2020byol,xie2021moby}. This ensures gradual evolution of $f_\xi$, stabilizing negative samples across iterations \citep{he2020moco}.

To generate synthetic hard negatives, we define the hardness of negative samples by their similarity to the query, measured through the logit values $\ell(\mathbf{n}_i) = \mathbf{q}^\top \cdot \mathbf{n}_i$. To identify the most challenging negatives, we order all negative features by decreasing similarity, \ie, $\mathcal{\hat{Q}} = \{\mathbf{n}_1, \mathbf{n}_2, \ldots, \mathbf{n}_K\}$ where $\ell(\mathbf{n}_i) > \ell(\mathbf{n}_j)$ for all $i < j$. The top-$N$ hardest negatives are then selected as $\mathcal{\hat{Q}}^N$ by truncating this ordered set. We define a general framework for synthetic negative generation where $\mathbf{s}_k^i$ represents the $k$-th synthetic negative from the $i$-th strategy. All synthetic negatives are $\ell_2$-normalized to ensure consistency with the representation space geometry. Following \citep{giakoumoglou2025synco,kalantidis2020mochi}, we implement six distinct transformation strategies:

\begin{equation}\label{eq:synthetic_strategies}
\mathbf{s}_k^i = \begin{cases}
\alpha_k \cdot \mathbf{q} + (1 - \alpha_k) \cdot \mathbf{n}_j, & i = 1 \\
\mathbf{n}_j + \beta_k \cdot (\mathbf{n}_j - \mathbf{q}), & i = 2 \\
\gamma_k \cdot \mathbf{n}_j + (1 - \gamma_k) \cdot \mathbf{n}_l, & i = 3  \\
\mathbf{n}_j + \mathcal{N}(\mathbf{0}, \sigma^2 \cdot \mathbf{I}), & i = 4  \\
\mathbf{n}_j + \delta \cdot \nabla_{\mathbf{n}_j} \text{sim}(\mathbf{q}, \mathbf{n}_j), & i = 5 \\
\mathbf{n}_j + \eta \cdot \text{sign}(\nabla_{\mathbf{n}_j} \text{sim}(\mathbf{q}, \mathbf{n}_j)), & i = 6
\end{cases}
\end{equation}

\noindent where $\mathbf{n}_j, \mathbf{n}_l \in \mathcal{\hat{Q}}^N$ are selected hard negatives, and $\text{sim}(\mathbf{q}, \mathbf{n}_j) = \mathbf{q}^\top \cdot \mathbf{n}_j$ represents the cosine similarity function. \textbf{(i)} \textit{Interpolated negatives} ($i=1$) create synthetic examples between the query and hard negatives, where $\alpha_k \in (0, 0.5)$ controls the balance between query and negative contributions. \textbf{(ii)} \textit{Extrapolated negatives} ($i=2$) explore directions beyond hard negatives, where $\beta_k \in (1, 1.5)$ determines the extrapolation distance. \textbf{(iii)} \textit{Mixup negatives} ($i=3$) combine pairs of hard negatives with mixing coefficient $\gamma_k \in (0, 1)$. \textbf{(iv)} \textit{Noise-injected negatives} ($i=4$) add controlled stochasticity with Gaussian noise ($\sigma = 0.01$). \textbf{(v)} \textit{Perturbed negatives} ($i=5$) modify hard negatives using gradient-based perturbations with $\delta = 0.01$. \textbf{(vi)} \textit{Adversarial negatives} ($i=6$) apply sign-based perturbations with strength $\eta = 0.01$.

The complete set of synthetic hard negatives is formed as $\mathcal{S} = \bigcup_{i=1}^{6} S^i$, where $S^i = \{\mathbf{s}_1^i, \mathbf{s}_2^i, \ldots\}$ contains all $|S^i|$ synthetic negatives generated by the $i$-th strategy. These synthetic negatives require memory size $\mathcal{O}(|\mathcal{S}| \cdot d)$, where $|\mathcal{S}| = \sum_{i=1}^{6} |S^i|\ll K$. We augment the memory queue's negative samples with synthetically generated hard negatives by calculating the denominator $Z$ that comprises contributions from both memory-based and synthetic negatives:

\begin{equation}\label{eq:memory_and_synthetic_negatives}
Z = \sum\limits_{\mathbf{n} \in \mathcal{Q}} \exp(\mathbf{q}^\top \cdot \mathbf{n} / \tau) + \sum\limits_{\mathbf{s} \in \mathcal{S}} \exp(\mathbf{q}^\top \cdot \mathbf{s} / \tau)
\end{equation}

\noindent where $\tau$ is the temperature parameter. We set $\tau=0.2$. Finally, we optimize the combined negative set using the InfoNCE loss function:

\begin{equation}\label{eq:loss_contrastive_with_synthetic_negatives}
    \mathcal{L}(\mathbf{q}, \mathbf{k}, \mathcal{Q}, \mathcal{S}) = -\log \frac{\exp(\mathbf{q}^\top \cdot \mathbf{k} / \tau)}{\exp(\mathbf{q}^\top \cdot \mathbf{k} / \tau) + Z}.
\end{equation}

\paragraph{\bf Relation to MoBY.}

When \textit{no} synthetic hard negatives are generated (\ie, $\mathcal{S} = \emptyset$), our method reduces to the standard InfoNCE loss used by MoBY \citep{xie2021moby} and MoCo-v3 \citep{chen2021mocov3} for vision transformers (\Cref{fig:moby}):

\begin{equation}\label{eq:loss_contrastive_moby}
\mathcal{L}_{\text{}}(\mathbf{q},\mathbf{k},\mathcal{Q}) =
-\log \frac{\exp(\mathbf{q}^\top \cdot \mathbf{k} / \tau )}{\exp(\mathbf{q}^\top \cdot \mathbf{k} / \tau ) + \sum\limits_{\mathbf{n} \in \mathcal{Q}} \exp(\mathbf{q}^\top \cdot \mathbf{n} / \tau)}.
\end{equation}

\paragraph{\bf Relation to BYOL.}

When we replace the InfoNCE loss with a mean squared error loss between the query $\mathbf{q}$ and key $\mathbf{k}$ representations, our method reduces to DINO \citep{caron2021dino} (\Cref{fig:dino}) without ``\textit{tricks}'' or to BYOL \citep{grill2020byol} (\Cref{fig:byol}):

\begin{equation}\label{eq:loss_ce_byol}
\mathcal{L}_{\text{MSE}}(\mathbf{q}, \mathbf{k}) = \frac{1}{2}\|\mathbf{q} - \mathbf{k}\|^2_2.
\end{equation}


\subsection{Implementation and Evaluation Protocols}
\label{sec:implementation}

\paragraph{\bf Architecture.}
We adopt ViT-S/16 (22M) and ViT-B/16 (86M) \citep{dosovitskiy2021vit,touvron2021deit} or Swin-T (28M) and Swin-S (50M) \citep{liu2021swin} as the backbone $f_\theta$. The projection ($g_\theta$) and prediction ($h_\theta$) heads are two-layer MLPs. Their hidden layers are 4096-dim with ReLU \citep{nair2010relu}, and outputs are 256-dim without ReLU. All MLP layers use BN \citep{ioffe2015batch}.

\paragraph{\bf Implementation details.}
We pretrain on ImageNet ILSVRC-2012 \citep{deng2009imagenet} and ImageNet-100 \citep{khosla2021supcon} without labels.Following MoBY~\citep{xie2021moby}, we train the model using AdamW~\citep{loshchilov2019adamw} with a batch size of 512, a base learning rate of $10^{-3}$, and a weight decay of $0.05$. Training spans 300 epochs. The target-network EMA parameter $m$ starts at $m_\text{start}=0.99$ and increases with cosine schedule to 1. We adopt BYOL augmentations~\citep{grill2020byol}. For synthetic negatives, we select the top $N=256$ negatives from the memory queue and generate $128$ synthetic hard negatives per anchor using six transformation strategies (\Cref{sec:vitamins}), totaling 768 synthetic negatives. Finally, we apply asymmetric drop path rates \citep{huang2016deep} of $0.2$ to the online encoder and $0.0$ to the target encoder, as in \citep{xie2021moby}. More implementation details in the supplementary material.

\paragraph{\bf Evaluation protocols.}
We follow standard self-supervised learning evaluation protocols to assess the quality of learned representations \citep{zhang2016colorization,he2020moco,chen2020simclr}. Three primary approaches are used: \textbf{(i)} \textit{linear probing evaluation}, where a linear classifier is trained on frozen features while keeping the backbone network fixed; \textbf{(ii)} \textit{full fine-tuning}, where all model parameters are updated on downstream tasks; and \textbf{(iii)} \textit{$k$-NN evaluation}, where the model's learned features are used to predict labels using a $k$-nearest neighbors classifier.

\section{Main Results}
\label{sec:main-results}

In this section, we present experimental results validating the effectiveness of \model for vision transformers, with implementation details and more results in the supplementary material.

\subsection{Linear Evaluation on ImageNet}
\label{sec:lineval}

We train a linear classifier on the frozen representation following standard protocols \citep{kornblith2019do,kolesnikov2019revisiting}, reporting top-1, top-5, and $k$-NN ($k=10$) accuracy in \Cref{tab:lineval-imagenet-deit,tab:lineval-imagenet-swin}. Our ViT-S surpasses I-JEPA ViT-B, and our ViT-B outperforms I-JEPA, V-JEPA, and iBOT despite their larger models. \model also surpasses MoBY (no synthetic negatives) and BYOL (no negatives) by a large margin on both linear and $k$-NN evaluation.

\begin{table}[!t]
\centering
\setlength{\tabcolsep}{1.2mm}
\caption{\textbf{Linear and $\boldsymbol{k}$-NN ViT classification on ImageNet.} 
Results show top-1 and top-5 accuracy and $k$-NN accuracy for models trained without multi-crop augmentation. 
\textbf{Symbols:} 
$^{\dagger}$ adapted from~\citep{chen2021mocov3};
$^{\ddagger}$ from~\citep{caron2021dino};
$^{\diamond}$ from~\citep{zhou2022ibot};
$^{\S}$ from~\citep{chen2023cae}. 
}
\label{tab:lineval-imagenet-deit}
\begin{tabular}{lccccc}
\toprule
\textbf{Method} & \textbf{Arch.} & \textbf{Ep.} & \textbf{Top-1} & \textbf{Top-5} & $\boldsymbol{k}$-NN\\
\midrule
Supervised~\citep{touvron2021deit} & ViT\text{-}S & 300 & 79.8 & -- & -- \\
Supervised~\citep{touvron2021deit} & ViT\text{-}B & 300 & 81.8 & -- & -- \\
\midrule
\multicolumn{6}{c}{\textit{Generative}} \\
MAE~\citep{he2021mae} & ViT\text{-}B & 1600 & 68.0 & -- & -- \\
SimMIM~\citep{xie2022simmim} & ViT\text{-}B & 800 & 56.7 & -- & -- \\
BeiT~\citep{bao2022beit}$^{\S}$ & ViT\text{-}S & 300 & 15.7 & -- & -- \\
CAE~\citep{chen2023cae}$^{\S}$ & ViT\text{-}S & 300 & 51.8 & -- & -- \\
CAE~\citep{chen2023cae} & ViT\text{-}B & 1600 & 70.4 & -- & -- \\
\midrule
\multicolumn{6}{c}{\textit{Joint Embedding Predictive Architectures}} \\
I-JEPA~\citep{assran2023ijepa} & ViT\text{-}B & 600 & 72.9 & -- & -- \\ 
V-JEPA~\citep{bardes2024vjepa} & ViT\text{-}L & 600 & 73.7 & -- & -- \\ 
LeJEPA~\citep{balestriero2025lejepa} & ViT\text{-}H & 100 & 77.1 & -- & -- \\ 
\midrule
\multicolumn{6}{c}{\textit{Joint Embedding Architectures}} \\
SwAV~\citep{caron2020swav}$^{\dagger}$ & ViT\text{-}S & 300 & 67.1 & -- & -- \\
BYOL~\citep{grill2020byol}$^{\dagger}$ & ViT\text{-}S & 300 & 71.0 & -- & -- \\
BYOL~\citep{grill2020byol} (repr.) & ViT\text{-}S & 300 & 70.3 & 91.0 & 62.5 \\
MoBY~\citep{xie2021moby} (repr.) & ViT\text{-}S & 300 & 72.3 & 88.3 & 64.3 \\
MoBY~\citep{xie2021moby} & ViT\text{-}S & 300 & 72.8 & -- & -- \\
MoCo-v3~\citep{chen2021mocov3}$^{\dagger}$ & ViT\text{-}S & 300 & 72.5 & -- & -- \\
MoCo-v3~\citep{chen2021mocov3}$^{\diamond}$ & ViT\text{-}B & 300 & 76.7 & -- & -- \\
DINO~\citep{caron2021dino}$^{\ddagger}$ & ViT\text{-}S & 300 & 72.5 & -- & 67.9 \\
DINO~\citep{caron2021dino} (repr.) & ViT\text{-}S & 300 & 72.3 & 93.5 & 66.5 \\
DINO~\citep{caron2021dino}$^{\diamond}$ & ViT\text{-}B & 200 & 76.0 & -- & 71.2 \\
iBOT~\citep{zhou2022ibot} & ViT\text{-}B & 200 & 76.0 & -- & 71.2 \\
\midrule
\model (ours) & ViT\text{-}S & 300 & \cellcolor{pearDarker!08}73.1 & \cellcolor{pearDarker!08}91.4 & \cellcolor{pearDarker!08}71.0 \\
\model (ours) & ViT\text{-}B & 300 & \cellcolor{pearDarker!08}\textbf{77.1} & \cellcolor{pearDarker!08}\textbf{94.4} & \cellcolor{pearDarker!08}\textbf{73.3} \\
\bottomrule
\end{tabular}
\end{table}
\begin{table}[!t]
\centering
\setlength{\tabcolsep}{1.2mm}
\caption{\textbf{Linear and $\boldsymbol{k}$-NN Swin classification on ImageNet.}
Results show top-1 and top-5 accuracy and $k$-NN accuracy
}
\label{tab:lineval-imagenet-swin}
\begin{tabular}{lccccc}
\toprule
\textbf{Method} & \textbf{Arch} & \textbf{Ep.} & \textbf{Top-1} & \textbf{Top-5} & $\boldsymbol{k}$-NN \\
\midrule
Supervised~\citep{liu2021swin} & Swin-T & 300 & 81.3 & -- & -- \\
Supervised~\citep{liu2021swin} & Swin-S & 300 & 83.0 & -- & -- \\
\midrule
\multicolumn{6}{c}{\textit{Generative}} \\
SiMIM~\citep{xie2022simmim} & Swin-T & 100 & 56.0 & -- & -- \\
\midrule
\multicolumn{6}{c}{\textit{Joint Embedding Architectures}} \\
BYOL~\citep{grill2020byol} (repr.) & Swin-T & 300 & 68.5 & 89.4 & 58.0 \\
MoBY~\citep{xie2021moby} & Swin-T & 300 & 75.0 & -- & -- \\
MoBY~\citep{xie2021moby} (repr.) & Swin-T & 300 & 74.7 & 92.7 & 67.8 \\
SMoG~\citep{pang2022smog} & Swin-T & 400 & 74.5 & -- & -- \\
\midrule
\model (ours) & Swin-T & 300 & \cellcolor{pearDarker!08}75.4 & \cellcolor{pearDarker!08}93.1 & \cellcolor{pearDarker!08}69.3 \\
\model (ours) & Swin-S & 300 & \cellcolor{pearDarker!08}\textbf{78.0} & \cellcolor{pearDarker!08}\textbf{95.6} & \cellcolor{pearDarker!08}\textbf{71.9} \\
\bottomrule
\end{tabular}
\end{table}

\subsection{Nearest Neighbor Retrieval}
\label{sec:nn_retrieval}

We further evaluate our representations on landmark retrieval and copy detection tasks to assess their effectiveness for matching and similarity search.

\paragraph{\bf Image retrieval.}
Following \citep{caron2021dino}, we consider the revisited \citep{radenovic2018revisited} Oxford and Paris datasets \citep{philbin2008lost}. We freeze the features and directly apply $k$-NN for retrieval. As shown in \Cref{tab:retrieval}, \model demonstrates competitive retrieval performance, with our smaller models achieving results comparable to or exceeding those of larger architectures.

\begin{table}[!t]
\centering
\setlength{\tabcolsep}{1.7mm}
\caption{\textbf{Image retrieval performance.} We report mAP on revisited Oxford ($\mathcal{R}$Ox) and Paris ($\mathcal{R}$Par) datasets. \textbf{Symbols}: $^{\lozenge}$ trained for more epochs with multi-crop augmentation.}
\label{tab:retrieval}
\begin{tabular}{lccccc}
\toprule
\multirow{2}{*}{\textbf{Method}} & \multirow{2}{*}{\textbf{Arch.}} & \multicolumn{2}{c}{\textbf{$\boldsymbol{\mathcal{R}}\text{Ox}$}} & \multicolumn{2}{c}{\textbf{$\boldsymbol{\mathcal{R}}\text{Par}$}} \\
\cmidrule(lr){3-4} \cmidrule(lr){5-6}
 & & M & H & M & H \\
\midrule
\textit{Supervised} & ViT-S & 33.5 & 8.9 & 63.0 & 37.2 \\
BYOL~\citep{grill2020byol} (repr.) & ViT-S & 23.8 & 5.4 & 52.2 & 20.5 \\
BYOL~\citep{grill2020byol} (repr.) & Swin-T & 24.1 & 4.1 & 49.7 & 18.7 \\
MoBY~\citep{xie2021moby} (repr.) & ViT-S & 32.4 & 6.8 & 61.9 & 25.2 \\
MoBY~\citep{xie2021moby} (repr.) & Swin-T & 32.4 & 7.3 & 61.5 & 24.4 \\
DINO~\citep{caron2021dino} (repr.) & ViT-S & 39.0 & 11.5 & 65.6 & 30.1 \\
DINO~\citep{caron2021dino}$^{\lozenge}$ & ViT-S & 37.2 & \textbf{13.7} & 63.1 & \textbf{34.4} \\
iBOT~\citep{zhou2022ibot}$^{\lozenge}$ & ViT-B & 36.6 & 13.0 & 61.5 & 34.1 \\
\midrule
\model (ours) & ViT-S & \cellcolor{pearDarker!08}\textbf{40.0} & \cellcolor{pearDarker!08}12.6 & \cellcolor{pearDarker!08}\textbf{66.8} & \cellcolor{pearDarker!08}31.3 \\
\model (ours) & Swin-T & \cellcolor{pearDarker!08}35.3 & \cellcolor{pearDarker!08}9.3 & \cellcolor{pearDarker!08}64.1 & \cellcolor{pearDarker!08}29.3 \\
\bottomrule
\end{tabular}
\end{table}

\paragraph{\bf Copy detection.}
We evaluate performance on copy detection following \citep{caron2021dino} protocol, reporting mean average precision on the ``strong'' subset of the Copydays dataset \citep{douze2009copydays}. As shown in \Cref{tab:copydays}, \model surpass DINO \citep{caron2021dino} on this task.

\begin{table}[!t]
\centering
\caption{\textbf{Copy detection.} We report the mAP performance in copy detection on Copydays ``strong'' subset~\citep{douze2009copydays}. All models use resolution $224^2$ and 1536 dimensions.}
\label{tab:copydays}
\begin{tabular}{lcc}
\toprule
\textbf{Method} & \textbf{Arch.} & \textbf{mAP} \\
\midrule
\textit{Supervised}~\citep{touvron2021deit} & ViT-B & 76.4 \\
DINO~\citep{caron2021dino} & ViT-B & 81.7 \\
DINO~\citep{caron2021dino} (repr.) & ViT-S & 78.4 \\
\midrule
\model (ours) & ViT-S & \cellcolor{pearDarker!08}79.7 \\
\model (ours) & ViT-B & \cellcolor{pearDarker!08}\textbf{82.0} \\
\bottomrule
\end{tabular}
\end{table}

\subsection{Discovering the Semantic Layout of Scenes}
\label{sec:discovering}

A remarkable property of self-supervised vision transformers, as shown by DINO \citep{caron2021dino,oquab2023dinov2}, is their ability to capture semantic scene structure without supervision. We evaluate this property through two complementary analyses: quantitative video segmentation performance (\Cref{tab:davis}) and qualitative visualization of learned attention patterns (\Cref{fig:attention_vit_comparison}).

\paragraph{\bf Video instance segmentation.}

Following \citep{jabri2020space}, we evaluate spatial coherence on DAVIS-2017 \citep{pont2017davis}, segmenting via nearest-neighbor matching between consecutive frames on frozen features without any training. \model achieves strong performance (\Cref{tab:davis}).

\begin{table}[!t]
\centering
\setlength{\tabcolsep}{0.4mm}
\caption{\textbf{DAVIS 2017 video object segmentation.} We report mean region similarity $\mathcal{J}_m$, mean contour-based accuracy $\mathcal{F}_m$, and their respective recall metrics $\mathcal{J}_r$ and $\mathcal{F}_r$. Image resolution is 480p.}
\label{tab:davis}
\begin{tabular}{lcccccc}
\toprule
\textbf{Method} & \textbf{Arch.} &
$\bm{(\mathcal{J} \& \mathcal{F})_m}$ &
$\bm{\mathcal{J}_m}$ &
$\bm{\mathcal{J}_r}$ &
$\bm{\mathcal{F}_m}$ &
$\bm{\mathcal{F}_r}$ \\
\midrule
BYOL~\citep{grill2020byol} (repr.) & ViT-S & 41.3 & 41.5 & 40.9 & 41.1 & 33.6 \\
BYOL~\citep{grill2020byol} (repr.) & Swin-T & 34.4 & 37.9 & 30.3 & 31.0 & 13.7 \\
MoBY~\citep{xie2021moby} (repr.) & ViT-S & 42.2 & 42.1 & 39.6 & 42.2 & 34.9 \\
MoBY~\citep{xie2021moby} (repr.) & Swin-T & 36.6 & 39.7 & 32.7 & 33.5 & 16.5 \\
DINO~\citep{caron2021dino} (repr.) & ViT-S & 39.1 & 40.3 & 37.4 & 39.2 & 34.8 \\
\midrule
DINO~\citep{caron2021dino}$^{\lozenge}$ & ViT-S & 61.8 & 60.2 & -- & 63.4 & -- \\
DINO~\citep{caron2021dino}$^{\lozenge}$ & ViT-B & \textbf{62.3} & \textbf{60.7} & -- & \textbf{63.9} & -- \\
iBOT~\citep{zhou2022ibot}$^{\lozenge}$ & ViT-B & 61.8 & 60.4 & -- & 63.2 & -- \\
\midrule
\model (ours) & ViT-S & \cellcolor{pearDarker!08}44.3 & \cellcolor{pearDarker!08}44.1 & \cellcolor{pearDarker!08}\textbf{41.8} & \cellcolor{pearDarker!08}44.5 & \cellcolor{pearDarker!08}\textbf{38.5} \\
\model (ours) & Swin-T & \cellcolor{pearDarker!08}37.6 & \cellcolor{pearDarker!08}40.5 & \cellcolor{pearDarker!08}32.1 & \cellcolor{pearDarker!08}34.6 & \cellcolor{pearDarker!08}17.0 \\
\bottomrule
\end{tabular}
\end{table}

\paragraph{\bf Visualizing attention mechanisms.}

Recent work \citep{caron2021dino,oquab2023dinov2} showed vision transformers segment objects and attend to meaningful regions without supervision, but whether this is exclusive to self-distillation or emerges more generally remains unclear. Following \citep{caron2021dino}, we visualize last-layer attention: \textbf{(i)} \texttt{[CLS]}-to-patch attention and \textbf{(ii)} patch self-attention capturing object boundaries. In \Cref{fig:teaser}, all methods separate foreground from background, but \model yields significantly sharper maps, capturing fine details like the bear's head and claws, the horse's body, and the dog's head and feet (\Cref{fig:teaser,fig:attention_vit_comparison}), without DINO's tricks \citep{caron2021dino} (\Cref{sec:related-work}). More in the supplementary material.

\begin{figure*}[!t]
    \centering
    \scriptsize
    \setlength{\tabcolsep}{2pt}
    \resizebox{\linewidth}{!}{
    \begin{tabular}{ccccccccc}
         & \multicolumn{4}{c}{\textbf{\texttt{CLS} Attention}} & \multicolumn{4}{c}{\textbf{Patch Attentions}} \\
         [2pt]
        \cmidrule(lr){2-5} \cmidrule(lr){6-9}
        \multirow{-1}{*}{Image} &
         &
         &
        Ours &
        Ours &
         &
         &
        Ours &
        Ours \\
        &
        \multirow{-2}{*}{MoBY}&
        \multirow{-2}{*}{BYOL} &
        ($odpr=0.1$) &
        ($odpr=0.2$) &
        \multirow{-2}{*}{MoBY} &
        \multirow{-2}{*}{BYOL} &
        $odpr=0.1$ &
        $odpr=0.2$\\
        [2pt]
        \includegraphics[width=0.10\linewidth]{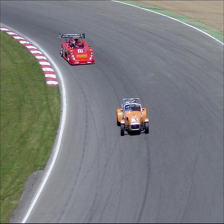} &
        \includegraphics[width=0.10\linewidth]{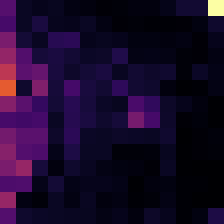} &
        \includegraphics[width=0.10\linewidth]{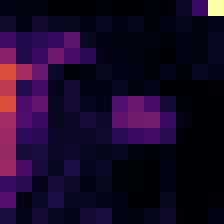} &
        \includegraphics[width=0.10\linewidth]{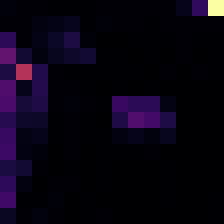} &
        \includegraphics[width=0.10\linewidth]{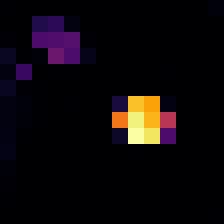} &
        \includegraphics[width=0.10\linewidth]{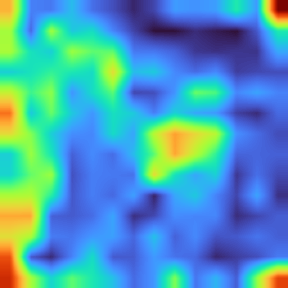} &
        \includegraphics[width=0.10\linewidth]{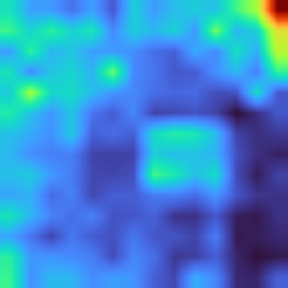} &
        \includegraphics[width=0.10\linewidth]{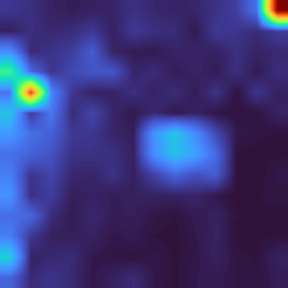} &
        \includegraphics[width=0.10\linewidth]{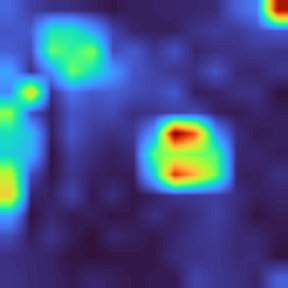} \\
        [2pt]
        \includegraphics[width=0.10\linewidth]{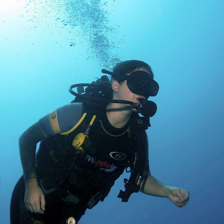} &
        \includegraphics[width=0.10\linewidth]{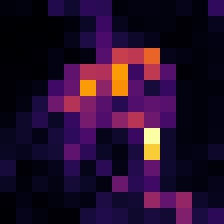} &
        \includegraphics[width=0.10\linewidth]{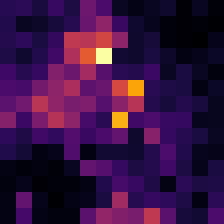} &
        \includegraphics[width=0.10\linewidth]{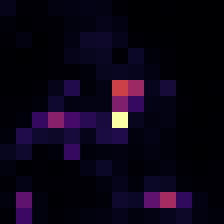} &
        \includegraphics[width=0.10\linewidth]{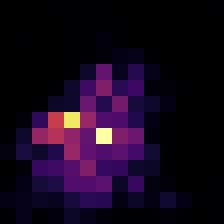} &
        \includegraphics[width=0.10\linewidth]{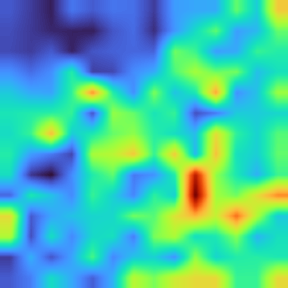} &
        \includegraphics[width=0.10\linewidth]{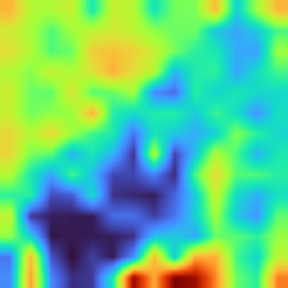} &
        \includegraphics[width=0.10\linewidth]{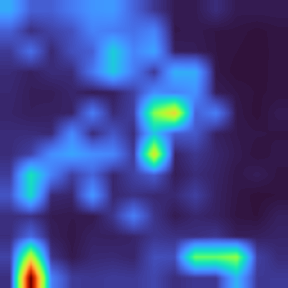} &
        \includegraphics[width=0.10\linewidth]{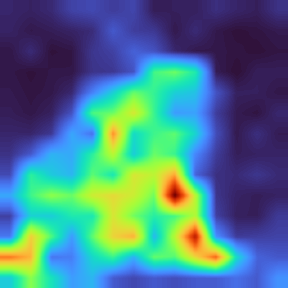} \\
        [2pt]
        \includegraphics[width=0.10\linewidth]{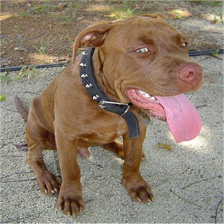} &
        \includegraphics[width=0.10\linewidth]{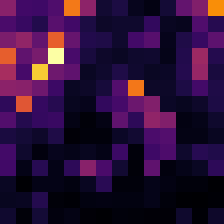} &
        \includegraphics[width=0.10\linewidth]{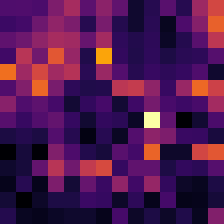} &
        \includegraphics[width=0.10\linewidth]{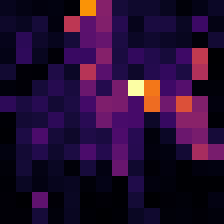} &
        \includegraphics[width=0.10\linewidth]{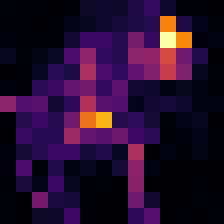} &
        \includegraphics[width=0.10\linewidth]{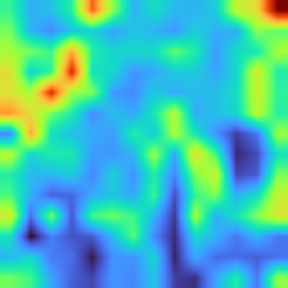} &
        \includegraphics[width=0.10\linewidth]{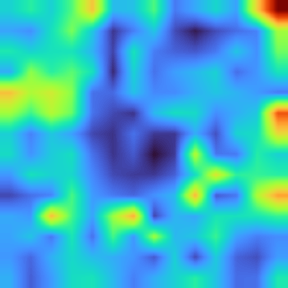} &
        \includegraphics[width=0.10\linewidth]{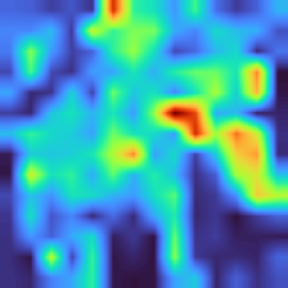} &
        \includegraphics[width=0.10\linewidth]{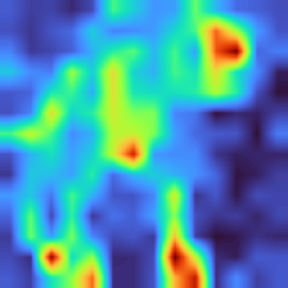} \\
        [2pt]
        \includegraphics[width=0.10\linewidth]{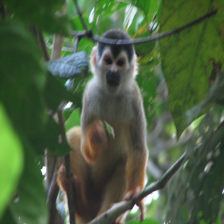} &
        \includegraphics[width=0.10\linewidth]{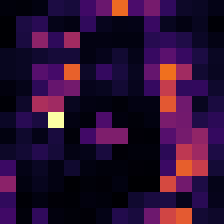} &
        \includegraphics[width=0.10\linewidth]{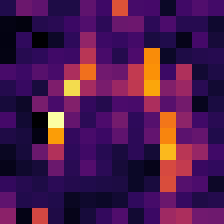} &
        \includegraphics[width=0.10\linewidth]{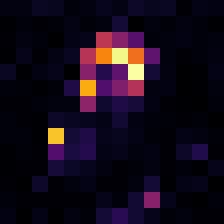} &
        \includegraphics[width=0.10\linewidth]{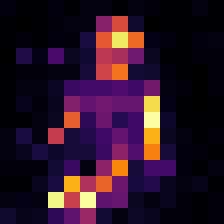} &
        \includegraphics[width=0.10\linewidth]{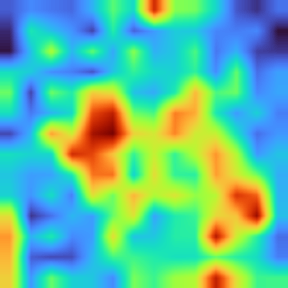} &
        \includegraphics[width=0.10\linewidth]{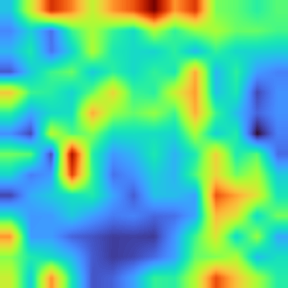} &
        \includegraphics[width=0.10\linewidth]{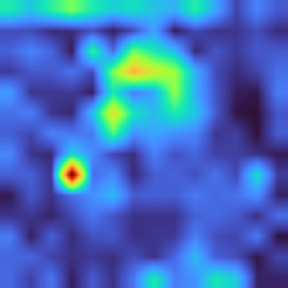} &
        \includegraphics[width=0.10\linewidth]{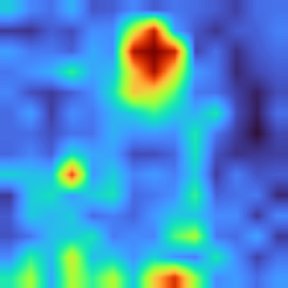} \\
    \end{tabular}}
    \caption{
    \textbf{ViT-S/16 attention visualization across self-supervised methods.} We show \texttt{CLS} attention (\textbf{left group}) and patch attention (\textbf{right group}) for MoBY, BYOL, and our proposed \model method for $odpr=0.1$ and $odpr=0.2$. More in the supplementary material.
    }
    \label{fig:attention_vit_comparison}
\end{figure*}

\subsection{Transfer Learning on Downstream Tasks}
\label{sec:transfer}

We evaluate \model transfer learning on dense prediction and classification tasks.

\begin{table}[!t]
\centering
\setlength{\tabcolsep}{0.9mm}
\caption{
\textbf{Transfer learning performance on dense prediction tasks.}
Object detection ($mAP^{bb}$) and instance segmentation ($mAP^{msk}$) are evaluated on COCO while semantic segmentation performance is reported as mIoU on ADE20K.
}
\label{tab:transfer-learning-dense}
\begin{tabular}{lcccc}
\toprule
\textbf{Method} & \textbf{Arch.} & $\mathbf{mAP^{bb}}$ & $\mathbf{mAP^{msk}}$ & $\mathbf{mIoU}$ \\
\midrule
\textit{Supervised}~\citep{touvron2021deit} & ViT-S & 46.2 & 40.1 & 44.5 \\
\textit{Supervised}~\citep{liu2021swin} & Swin-T & 48.1 & 41.7 & 44.5 \\
iBOT~\citep{zhou2022ibot} & ViT-S & 49.4 & 42.6 & 45.4 \\
MoBY~\citep{xie2021moby} & Swin-T & 48.1 & 41.5 & 44.1 \\
DINO~\citep{caron2021dino} (repr.) & ViT-S & 41.5 & 41.3 & 45.0 \\
\midrule
\model (ours) & ViT-S & \cellcolor{pearDarker!08}\textbf{49.9} & \cellcolor{pearDarker!08}\textbf{42.8} & \cellcolor{pearDarker!08}\textbf{46.0} \\
\model (ours) & Swin-T & \cellcolor{pearDarker!08}48.9 & \cellcolor{pearDarker!08}42.2 & \cellcolor{pearDarker!08}45.2 \\
\bottomrule
\end{tabular}
\end{table}\textbf{}

\paragraph{\bf Detection and segmentation on COCO.}
We evaluate on the COCO dataset \citep{li2015coco} using Cascade Mask R-CNN \citep{cai2019cascadercnn}, which simultaneously predicts bounding boxes and instance masks. As reported in \Cref{tab:transfer-learning-dense}, our method consistently improves over strong self-supervised baselines across both architectures.

\paragraph{\bf Semantic segmentation on ADE20K.}
We further evaluate representation quality on semantic segmentation, a dense pixel-level classification task, using ADE20K \citep{zhou2018ade20k} with UPerNet \citep{xiao2018upernet}. As shown in \Cref{tab:transfer-learning-dense}, our method achieves the best mIoU across transformer architectures.

\paragraph{\bf Transfer learning.}
Finally, we test whether \model's ImageNet features are generic or ImageNet-specific, performing linear evaluation and fine-tuning on the classification tasks of \citep{kolesnikov2019revisiting,kornblith2019do} (\Cref{tab:transfer-learning-linprobe,tab:transfer-learning-ft}). In linear probing, our method wins on 8/11 datasets with ViT-S and 9/11 with Swin-T (see the supplementary material); in fine-tuning, it outperforms both baselines on all 5 datasets with both architectures. Representations transfer effectively to small images (CIFAR-10/100 \citep{krizhevsky2009cifar}), fine-grained recognition (Flowers-102 \citep{nilsback2008flowers102}, Cars \citep{krause2013cars}), landscapes (SUN397 \citep{xiao2010sun397}), and textures (DTD \citep{cimpoi2014dtd}). 

\begin{table}[!t]
\caption{\textbf{Linear probing performance on various downstream classification datasets.} Results show top-1 accuracy (in \%) with frozen weights except for the final fully-connected layer. \textbf{Abbreviations}: C: CIFAR, Flwr: Flowers-102, F: Food.}
\label{tab:transfer-learning-linprobe}
\centering
\begin{tabular}{lccccc}
\toprule
\textbf{Method} & \textbf{C\textsubscript{10}} & \textbf{C\textsubscript{100}} & \textbf{Flwr} & \textbf{Pets} & \textbf{F\textsubscript{101}} \\
\midrule
\multicolumn{6}{c}{ViT-S} \\
BYOL~\citep{grill2020byol} (repr.) & 90.5 & 74.2 & \textbf{87.7} & 85.1 & 73.3 \\
MoBY~\citep{xie2021moby} (repr.)  & 88.9 & 73.0 & 56.8 & 80.8 & 69.7 \\
\model (ours) & \cellcolor{pearDarker!08}\textbf{92.1} & \cellcolor{pearDarker!08}\textbf{79.7} & \cellcolor{pearDarker!08}72.6 & \cellcolor{pearDarker!08}\textbf{86.1} & \cellcolor{pearDarker!08}\textbf{75.0} \\
\midrule
\multicolumn{6}{c}{Swin-T} \\
BYOL~\citep{grill2020byol} (repr.) & 88.6 & 72.2 & 83.8 & 83.0 & 73.7 \\
MoBY~\citep{xie2021moby} (repr.)  & 90.6 & 76.5 & \textbf{90.3} & 88.2 & 78.8 \\
\model (ours) & \cellcolor{pearDarker!08}\textbf{91.4} & \cellcolor{pearDarker!08}\textbf{77.7} & \cellcolor{pearDarker!08}89.5 & \cellcolor{pearDarker!08}\textbf{88.5} & \cellcolor{pearDarker!08}\textbf{79.8} \\
\bottomrule
\end{tabular}
\vspace{-0.2cm}
\end{table}
\begin{table}[!t]
\caption{\textbf{End-to-end finetuning performance on various downstream classification tasks.} Results show top-1 accuracy (in \%) with all parameters updated during training. \textbf{Abbreviations}: C: CIFAR, Flwr: Flowers-102, F: Food.}
\label{tab:transfer-learning-ft}
\centering
\begin{tabular}{lccccc}
\toprule
\textbf{Method} & \textbf{C\textsubscript{10}} & \textbf{C\textsubscript{100}} & \textbf{Flwr} & \textbf{Pets} & \textbf{F\textsubscript{101}} \\
\midrule
\multicolumn{6}{c}{ViT-S} \\
BYOL~\citep{grill2020byol} (repr.) & 86.3 & 62.4 & \textbf{87.7} & 85.1 & 73.5 \\
MoBY~\citep{xie2021moby} (repr.)  & 75.2 & 80.3 & 66.0 & 82.3 & 71.0 \\
\model (ours) & \cellcolor{pearDarker!08}\textbf{96.8} & \cellcolor{pearDarker!08}\textbf{83.1} & \cellcolor{pearDarker!08}88.3 & \cellcolor{pearDarker!08}\textbf{87.2} & \cellcolor{pearDarker!08}\textbf{85.7} \\
\midrule
\multicolumn{6}{c}{Swin-T} \\
BYOL~\citep{grill2020byol} (repr.) & 89.2 & 64.9 & 83.8 & 83.0 & 74.0 \\
MoBY~\citep{xie2021moby} (repr.)  & 97.3 & 84.8 & \textbf{90.3} & 88.2 & 79.8 \\
\model (ours) & \cellcolor{pearDarker!08}\textbf{97.6} & \cellcolor{pearDarker!08}\textbf{85.8} & \cellcolor{pearDarker!08}91.2 & \cellcolor{pearDarker!08}\textbf{89.5} & \cellcolor{pearDarker!08}\textbf{90.3} \\
\bottomrule
\end{tabular}
\vspace{-0.2cm}
\end{table}

\section{Ablation Study of \model}
\label{sec:ablation}

We conduct ablations to analyze synthetic negative strategies, regularization techniques, and hyperparameters. More ablations in the supplementary material.

\begin{table}[!t]
\centering
\caption{\textbf{Ablation study on synthetic negative strategies on ImageNet.} Each strategy generates 128 synthetic negatives. We pretrain for 100 epochs and report top-1 accuracy (\%). We {\fboxsep 1pt\colorbox{gray!10}{highlight}} the default hyperparameter.}
\label{tab:synth_neg_ablation}
\begin{tabular}{cccccccc}
\toprule
$\boldsymbol{S^1}$ & $\boldsymbol{S^2}$ & $\boldsymbol{S^3}$ & $\boldsymbol{S^4}$ & $\boldsymbol{S^5}$ & $\boldsymbol{S^6}$ & \textbf{\textit{ViT-S}} & \textbf{\textit{Swin-T}} \\
\midrule
\ding{55} & \ding{55} & \ding{55} & \ding{55} & \ding{55} & \ding{55} & 69.2 & 70.9 \\ 
\cellcolor{gray!10}\checkmark & \ding{55} & \ding{55} & \ding{55} & \ding{55} & \ding{55} & 69.5 & 71.2 \\ 
\ding{55} & \cellcolor{gray!10}\checkmark & \ding{55} & \ding{55} & \ding{55} & \ding{55} & 69.4 & 71.1 \\ 
\ding{55} & \ding{55} & \cellcolor{gray!10}\checkmark & \ding{55} & \ding{55} & \ding{55} & 69.6 & 71.3 \\ 
\ding{55} & \ding{55} & \ding{55} & \cellcolor{gray!10}\checkmark & \ding{55} & \ding{55} & 69.4 & 71.1 \\ 
\ding{55} & \ding{55} & \ding{55} & \ding{55} & \cellcolor{gray!10}\checkmark & \ding{55} & 69.3 & 71.0 \\ 
\ding{55} & \ding{55} & \ding{55} & \ding{55} & \ding{55} & \cellcolor{gray!10}\checkmark & 69.3 & 71.0 \\ 
\midrule
\cellcolor{gray!10}\checkmark & 
\cellcolor{gray!10}\checkmark &
\cellcolor{gray!10}\checkmark &
\cellcolor{gray!10}\checkmark &
\cellcolor{gray!10}\checkmark &
\cellcolor{gray!10}\checkmark &
\textbf{70.0} & \textbf{71.6} \\
\bottomrule
\end{tabular}
\end{table}

\begin{table}[!t]
\centering
\caption{\textbf{Ablation study on the drop path rates on ImageNet.} We pretrain and report top-1 accuracy (\%). We {\fboxsep 1pt\colorbox{gray!10}{highlight}} the default hyperparameter.}
\label{tab:ablation_dpr}
\begin{tabular}{ccccc}
\toprule
\textbf{\emph{Online} dpr} & \textbf{\emph{Target} dpr} & \textbf{Epochs} & \textbf{ViT-S} & \textbf{Swin-T} \\
\midrule
0.0  & 0.0 & 100 & 68.3 & 70.0 \\
0.1  & 0.1 & 100 & 68.1 & 69.8 \\
0.05 & 0.0 & 100 & 69.7 & 71.4 \\
0.1  & 0.0 & 100 & 69.8 & 71.5 \\
\cellcolor{gray!10}0.2 & \cellcolor{gray!10}0.0 & 100 & \textbf{70.0} & \textbf{71.6} \\
\midrule
0.1  & 0.0 & 300 & 72.5 & 74.7 \\
\cellcolor{gray!10}0.2 & \cellcolor{gray!10}0.0 & 300 & \textbf{73.1} & \textbf{75.4} \\
\bottomrule
\end{tabular}
\end{table}

\begin{table}[!t]
\centering
\caption{\textbf{Ablation study on applying MoCo-v3 and SynCo \textit{tricks} on ImageNet.} We pretrain for 300 epochs and report top-1 accuracy (\%).  We {\fboxsep 1pt\colorbox{gray!10}{highlight}} the default hyperparameter.}
\label{tab:ablation_mocov3}
\begin{tabular}{cccc}
\toprule
\textbf{Fixed Patch Embed} & \textbf{Cooldown} & \textbf{ViT-S} & \textbf{Swin-T} \\
\midrule
\checkmark & \ding{55} & 72.0 & 73.6 \\
\ding{55} & \ding{55} & 72.2 & 74.1 \\
\midrule
\cellcolor{gray!10}\ding{55} & \cellcolor{gray!10}\checkmark & \textbf{73.1} & \textbf{75.4} \\
\bottomrule
\end{tabular}
\end{table}

\paragraph{\bf Synthetic hard negatives strategies.}

We perform ablation studies on combinations of synthetic negative transformation strategies. \Cref{tab:synth_neg_ablation} shows that combining all six types ($S^1$–$S^6$) yields the highest performance. Without synthetic negatives, the baseline improves by \textbf{+0.8\%} and \textbf{+0.7\%} when all strategies are applied. While individual strategies vary in effectiveness ($S^3$ most impactful, then $S^1$), their combination provides complementary benefits exceeding individual contributions, validating that diverse synthetic negatives improve representations.

\begin{figure*}[!t]
\centering
\includegraphics[width=0.24\textwidth]{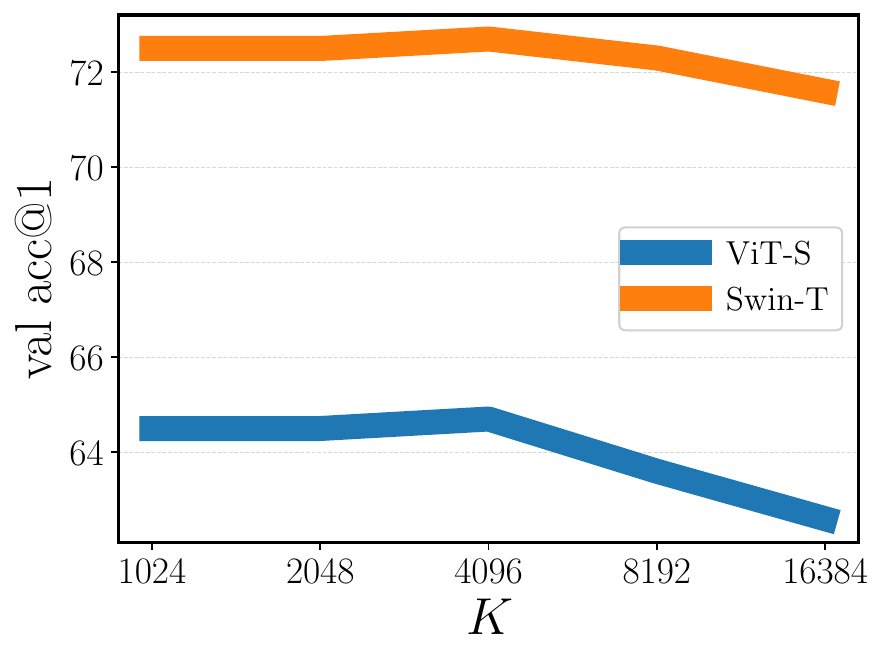}
\hfill
\includegraphics[width=0.24\textwidth]{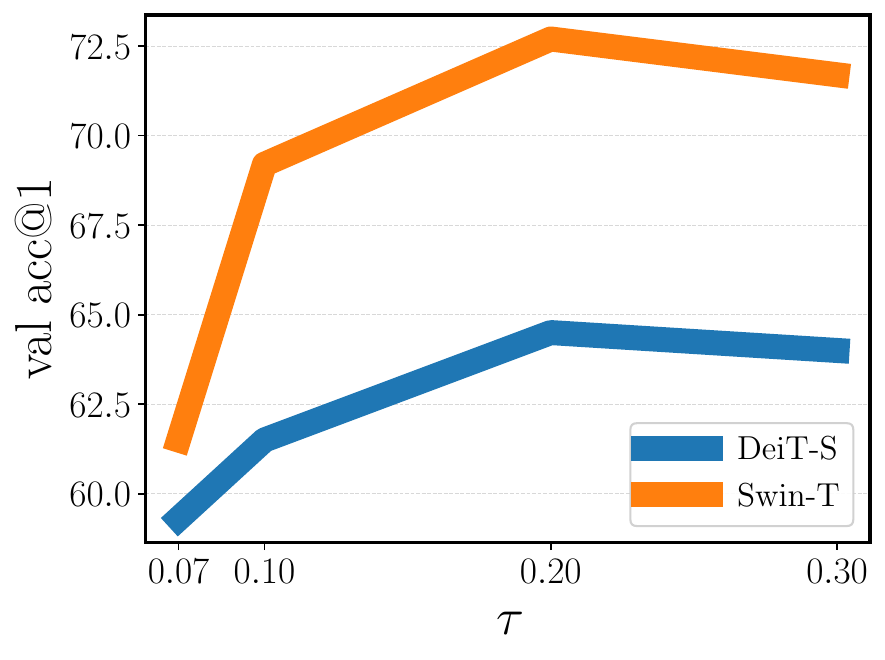}
\hfill
\includegraphics[width=0.24\textwidth]{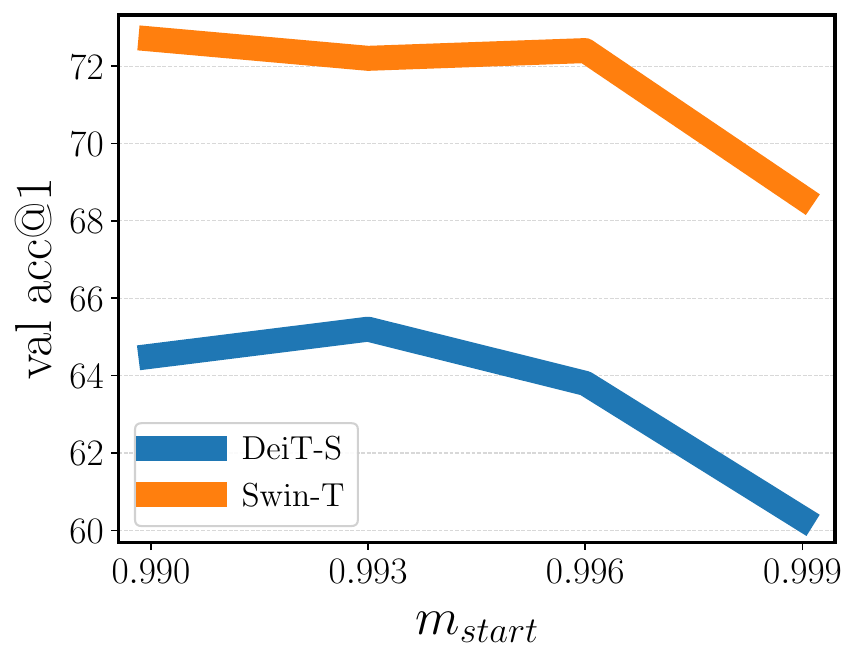}
\hfill
\includegraphics[width=0.24\textwidth]{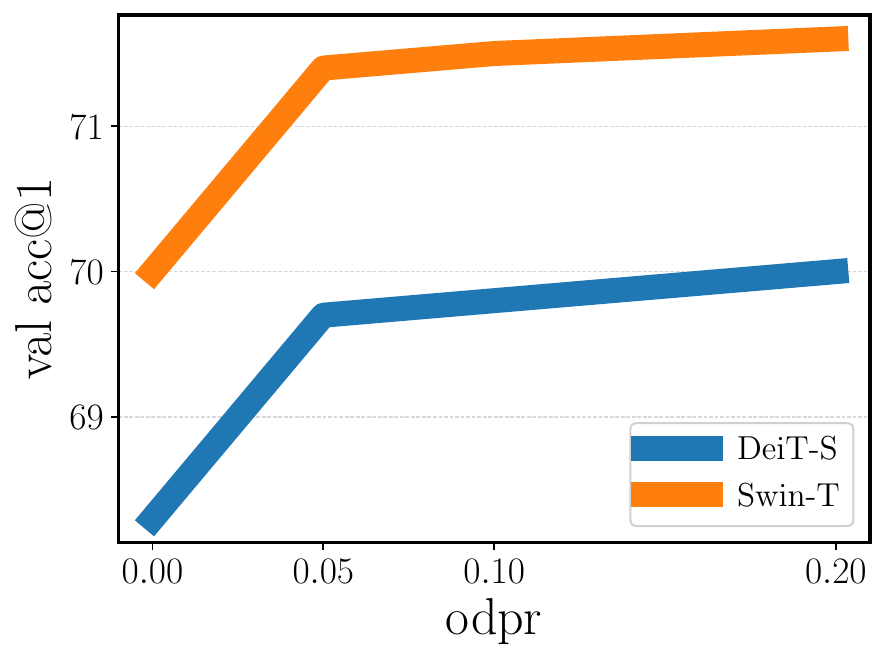}
\caption{\textbf{Ablation studies of \model on ImageNet-100.} We pretrain for 100 epochs and report top-1 accuracy (\%) using ViT-S and Swin-T architectures. (\textbf{left}): queue size $K$; (\textbf{second from left}): temperature $\tau$; (\textbf{second from right}): momentum $m_{\text{start}}$; (\textbf{right}): \textit{online} drop path rate (ImageNet here, see \cref{tab:ablation_dpr}). \textbf{Default hyperparameters}: $K=4096$, $\tau=0.2$, $m_{\text{start}}=0.99$, \textit{online} drop path rate $=0.2$.}
\label{fig:ablation_hyperparams}
\end{figure*}

\paragraph{\bf Drop path regularization.}

We investigate the effect of asymmetric drop path on \textit{online} and \textit{target} encoders. \Cref{tab:ablation_dpr} shows that higher drop path rates for the online encoder with no dropout for the target encoder yields optimal results. This asymmetric configuration outperforms both no regularization and symmetric drop path rates, with the effectiveness likely stemming from encouraging the online encoder to learn more robust representations while maintaining stability in the target encoder. \Cref{tab:ablation_dpr} also shows how our method benefits from extended pretraining.

\paragraph{\bf Tricks of MoCo-v3 and SynCo.}

We evaluate the necessity of implementation \textit{tricks} from MoCo-v3 \citep{chen2021mocov3} and SynCo \citep{giakoumoglou2025synco}. \Cref{tab:ablation_mocov3} shows that fixed patch embeddings from MoCo-v3 are unnecessary when using our approach, while SynCo's cooldown strategy (disabling synthetic negatives for the last 100 epochs) achieves optimal results. This cooldown approach (along warmup) has proven effective for both convnets \citep{giakoumoglou2025synco} and vision transformers (ours). Synthetics without warmup hurt because early representations are not yet semantically organized, while synthetics without cooldown destabilize convergence. A full ablation of the three phases is provided in the supplementary material.

\paragraph{\bf Other hyperparameters.}

We study the robustness of our approach across different contrastive hyperparameter settings to demonstrate seamless integration with existing frameworks. \Cref{fig:ablation_hyperparams} shows that performance remains stable across a wide range of queue sizes, temperatures, and momentum values using default hyperparameters from MoBY. These findings confirm that synthetic negatives can be readily adopted in existing contrastive learning pipelines \textit{without} requiring architectural modifications, extensive hyperparameter re-tuning, or additional computational overhead during the hyperparameter search process.

\section{Conclusion}
\label{sec:conclusion}

We demonstrate that synthetic hard negatives significantly improve vision transformer representations in self-supervised contrastive learning. Emergent semantic segmentation properties, previously considered exclusive to self-distillation methods like DINO \citep{caron2021dino}, naturally arise in contrastive learning and are further strengthened through synthetic negative generation. \model achieves five improvements over standard contrastive baselines: \textbf{(i)} improved ImageNet linear accuracy, \textbf{(ii)} strong $k$-NN performance indicating high-quality features, \textbf{(iii)} improved downstream performance across diverse settings, \textbf{(iv)} sharper attention maps with better object boundary alignment, and \textbf{(v)} strong video object segmentation despite no video training. These gains are achieved without DINO's complex procedures (centering, sharpening, multi-crop, extended schedules); under identical regimes (300~epochs, no multi-crop), \model outperforms our reproduced DINO across \textit{all} tasks: linear probing, $k$-NN, image retrieval, copy detection, video segmentation, and dense prediction (\Cref{tab:lineval-imagenet-deit,tab:retrieval,tab:copydays,tab:davis,tab:transfer-learning-dense}). See the supplementary material for scope, fair comparison (including DINO), and broader gains. Overall, our results challenge the focus on self-distillation and generative approaches: contrastive learning with high-quality negatives remains a simple yet powerful alternative that integrates with any InfoNCE-based method \citep{oord2019cpc}, generalizes across architectures, and incurs minimal overhead. We hope this encourages renewed interest in contrastive learning.

\section*{Acknowledgments}

We acknowledge the computational resources and support provided by the Imperial College Research Computing Service (\url{http://doi.org/10.14469/hpc/2232}), which enabled our experiments.

{
    \small
    \bibliographystyle{ieeenat_fullname}
    \bibliography{main}

@misc{cai2019cascadercnn,
	title        = {Cascade R-CNN: High Quality Object Detection and Instance Segmentation},
	author       = {Zhaowei Cai and Nuno Vasconcelos},
	year         = 2019,
	url          = {https://arxiv.org/abs/1906.09756},
	eprint       = {1906.09756},
	archiveprefix = {arXiv},
	primaryclass = {cs.CV}
}

@inproceedings{douze2009copydays,
	title        = {Evaluation of GIST descriptors for web-scale image search},
	author       = {Douze, Matthijs and J\'{e}gou, Herv\'{e} and Sandhawalia, Harsimrat and Amsaleg, Laurent and Schmid, Cordelia},
	year         = 2009,
	booktitle    = {Proceedings of the ACM International Conference on Image and Video Retrieval},
	location     = {Santorini, Fira, Greece},
	publisher    = {Association for Computing Machinery},
	address      = {New York, NY, USA},
	series       = {CIVR '09},
	doi          = {10.1145/1646396.1646421},
	isbn         = 9781605584805,
	url          = {https://doi.org/10.1145/1646396.1646421},
	articleno    = 19,
	numpages     = 8
}

@misc{xiao2018upernet,
	title        = {Unified Perceptual Parsing for Scene Understanding},
	author       = {Tete Xiao and Yingcheng Liu and Bolei Zhou and Yuning Jiang and Jian Sun},
	year         = 2018,
	url          = {https://arxiv.org/abs/1807.10221},
	eprint       = {1807.10221},
	archiveprefix = {arXiv},
	primaryclass = {cs.CV}
}

@misc{assran2023ijepa,
	title        = {Self-Supervised Learning from Images with a Joint-Embedding Predictive Architecture},
	author       = {Mahmoud Assran and Quentin Duval and Ishan Misra and Piotr Bojanowski and Pascal Vincent and Michael Rabbat and Yann LeCun and Nicolas Ballas},
	year         = 2023,
	url          = {https://arxiv.org/abs/2301.08243},
	eprint       = {2301.08243},
	archiveprefix = {arXiv},
	primaryclass = {cs.CV}
}

@misc{ali2024overfittingrobustnessquantityquality,
	title        = {From Overfitting to Robustness: Quantity, Quality, and Variety Oriented Negative Sample Selection in Graph Contrastive Learning},
	author       = {Adnan Ali and Jinlong Li and Huanhuan Chen and Ali Kashif Bashir},
	year         = 2024,
	url          = {https://arxiv.org/abs/2406.15044},
	eprint       = {2406.15044},
	archiveprefix = {arXiv},
	primaryclass = {cs.LG}
}

@misc{simeoni2025dinov3,
	title        = {DINOv3},
	author       = {Oriane Siméoni and Huy V. Vo and Maximilian Seitzer and Federico Baldassarre and Maxime Oquab and Cijo Jose and Vasil Khalidov and Marc Szafraniec and Seungeun Yi and Michaël Ramamonjisoa and Francisco Massa and Daniel Haziza and Luca Wehrstedt and Jianyuan Wang and Timothée Darcet and Théo Moutakanni and Leonel Sentana and Claire Roberts and Andrea Vedaldi and Jamie Tolan and John Brandt and Camille Couprie and Julien Mairal and Hervé Jégou and Patrick Labatut and Piotr Bojanowski},
	year         = 2025,
	url          = {https://arxiv.org/abs/2508.10104},
	eprint       = {2508.10104},
	archiveprefix = {arXiv},
	primaryclass = {cs.CV}
}

@misc{giakoumoglou2024review,
	title        = {A Review on Discriminative Self-supervised Learning Methods in Computer Vision},
	author       = {Nikolaos Giakoumoglou and Tania Stathaki and Athanasios Gkelias},
	year         = 2025,
	url          = {https://arxiv.org/abs/2405.04969},
	eprint       = {2405.04969},
	archiveprefix = {arXiv},
	primaryclass = {cs.CV}
}

@article{bardes2024vjepa,
	title        = {Revisiting Feature Prediction for Learning Visual Representations from Video},
	author       = {Bardes, Adrien and Garrido, Quentin and Ponce, Jean and Rabbat, Michael and LeCun, Yann and Assran, Mahmoud and Ballas, Nicolas},
	year         = 2024,
	journal      = {arXiv preprint arXiv:2404.08471}
}

@misc{pont2017davis,
	title        = {The 2017 DAVIS Challenge on Video Object Segmentation},
	author       = {Jordi Pont-Tuset and Federico Perazzi and Sergi Caelles and Pablo Arbeláez and Alex Sorkine-Hornung and Luc Van Gool},
	year         = 2018,
	url          = {https://arxiv.org/abs/1704.00675},
	eprint       = {1704.00675},
	archiveprefix = {arXiv},
	primaryclass = {cs.CV}
}

@inproceedings{jabri2020space,
	title        = {Space-Time Correspondence as a Contrastive Random Walk},
	author       = {Jabri, Allan and Owens, Andrew and Efros, Alexei},
	year         = 2020,
	booktitle    = {Advances in Neural Information Processing Systems},
	publisher    = {Curran Associates, Inc.},
	volume       = 33,
	pages        = {19545--19560},
	url          = {https://proceedings.neurips.cc/paper_files/paper/2020/file/e2ef524fbf3d9fe611d5a8e90fefdc9c-Paper.pdf},
	editor       = {H. Larochelle and M. Ranzato and R. Hadsell and M.F. Balcan and H. Lin}
}

@misc{radenovic2018revisited,
	title        = {Revisiting Oxford and Paris: Large-Scale Image Retrieval Benchmarking},
	author       = {Filip Radenović and Ahmet Iscen and Giorgos Tolias and Yannis Avrithis and Ondřej Chum},
	year         = 2018,
	url          = {https://arxiv.org/abs/1803.11285},
	eprint       = {1803.11285},
	archiveprefix = {arXiv},
	primaryclass = {cs.CV}
}

@inproceedings{philbin2008lost,
	title        = {Lost in quantization: Improving particular object retrieval in large scale image databases},
	author       = {Philbin, James and Chum, Ondrej and Isard, Michael and Sivic, Josef and Zisserman, Andrew},
	year         = 2008,
	booktitle    = {2008 IEEE Conference on Computer Vision and Pattern Recognition},
	volume       = {},
	number       = {},
	pages        = {1--8},
	doi          = {10.1109/CVPR.2008.4587635}
}

@article{bommasani2021foundationmodels,
	title        = {On the Opportunities and Risks of Foundation Models},
	author       = {Rishi Bommasani and Drew A. Hudson and Ehsan Adeli and Russ Altman and Simran Arora and Sydney von Arx and Michael S. Bernstein and Jeannette Bohg and Antoine Bosselut and Emma Brunskill and Erik Brynjolfsson and S. Buch and Dallas Card and Rodrigo Castellon and Niladri S. Chatterji and Annie S. Chen and Kathleen A. Creel and Jared Davis and Dora Demszky and Chris Donahue and Moussa Doumbouya and Esin Durmus and Stefano Ermon and John Etchemendy and Kawin Ethayarajh and Li Fei-Fei and Chelsea Finn and Trevor Gale and Lauren E. Gillespie and Karan Goel and Noah D. Goodman and Shelby Grossman and Neel Guha and Tatsunori Hashimoto and Peter Henderson and John Hewitt and Daniel E. Ho and Jenny Hong and Kyle Hsu and Jing Huang and Thomas F. Icard and Saahil Jain and Dan Jurafsky and Pratyusha Kalluri and Siddharth Karamcheti and Geoff Keeling and Fereshte Khani and O. Khattab and Pang Wei Koh and Mark S. Krass and Ranjay Krishna and Rohith Kuditipudi and Ananya Kumar and Faisal Ladhak and Mina Lee and Tony Lee and Jure Leskovec and Isabelle Levent and Xiang Lisa Li and Xuechen Li and Tengyu Ma and Ali Malik and Christopher D. Manning and Suvir P. Mirchandani and Eric Mitchell and Zanele Munyikwa and Suraj Nair and Avanika Narayan and Deepak Narayanan and Benjamin Newman and Allen Nie and Juan Carlos Niebles and Hamed Nilforoshan and J. F. Nyarko and Giray Ogut and Laurel Orr and Isabel Papadimitriou and Joon Sung Park and Chris Piech and Eva Portelance and Christopher Potts and Aditi Raghunathan and Robert Reich and Hongyu Ren and Frieda Rong and Yusuf H. Roohani and Camilo Ruiz and Jack Ryan and Christopher R'e and Dorsa Sadigh and Shiori Sagawa and Keshav Santhanam and Andy Shih and Krishna Parasuram Srinivasan and Alex Tamkin and Rohan Taori and Armin W. Thomas and Florian Tram{\`e}r and Rose E. Wang and William Wang and Bohan Wu and Jiajun Wu and Yuhuai Wu and Sang Michael Xie and Michihiro Yasunaga and Jiaxuan You and Matei A. Zaharia and Michael Zhang and Tianyi Zhang and Xikun Zhang and Yuhui Zhang and Lucia Zheng and Kaitlyn Zhou and Percy Liang},
	year         = 2021,
	journal      = {ArXiv},
	url          = {https://crfm.stanford.edu/assets/report.pdf}
}

@inproceedings{nilsback2008flowers102,
	title        = {Automated Flower Classification over a Large Number of Classes},
	author       = {Nilsback, Maria-Elena and Zisserman, Andrew},
	year         = 2008,
	booktitle    = {2008 Sixth Indian Conference on Computer Vision, Graphics and Image Processing},
	volume       = {},
	number       = {},
	pages        = {722--729},
	doi          = {10.1109/ICVGIP.2008.47}
}

@article{deng2009imagenet,
	title        = {ImageNet: A large-scale hierarchical image database},
	author       = {Jia Deng and Wei Dong and Richard Socher and Li-Jia Li and K. Li and Li Fei-Fei},
	year         = 2009,
	journal      = {2009 IEEE Conference on Computer Vision and Pattern Recognition},
	pages        = {248--255},
	url          = {https://api.semanticscholar.org/CorpusID:57246310}
}

@article{krizhevsky2009cifar,
	title        = {Learning Multiple Layers of Features from Tiny Images},
	author       = {Krizhevsky, Alex},
	year         = 2009,
	pages        = {32--33},
	url          = {https://www.cs.toronto.edu/~kriz/learning-features-2009-TR.pdf}
}

@inproceedings{xiao2010sun397,
	title        = {SUN database: Large-scale scene recognition from abbey to zoo},
	author       = {Xiao, Jianxiong and Hays, James and Ehinger, Krista A. and Oliva, Aude and Torralba, Antonio},
	year         = 2010,
	booktitle    = {2010 IEEE Computer Society Conference on Computer Vision and Pattern Recognition},
	volume       = {},
	number       = {},
	pages        = {3485--3492},
	doi          = {10.1109/CVPR.2010.5539970}
}

@inproceedings{krause2013cars,
	title        = {3D Object Representations for Fine-Grained Categorization},
	author       = {Krause, Jonathan and Stark, Michael and Deng, Jia and Fei-Fei, Li},
	year         = 2013,
	booktitle    = {2013 IEEE International Conference on Computer Vision Workshops},
	volume       = {},
	number       = {},
	pages        = {554--561},
	doi          = {10.1109/ICCVW.2013.77}
}

@inproceedings{cimpoi2014dtd,
	title        = {Describing Textures in the Wild},
	author       = {Cimpoi, Mircea and Maji, Subhransu and Kokkinos, Iasonas and Mohamed, Sammy and Vedaldi, Andrea},
	year         = 2014,
	booktitle    = {2014 IEEE Conference on Computer Vision and Pattern Recognition},
	volume       = {},
	number       = {},
	pages        = {3606--3613},
	doi          = {10.1109/CVPR.2014.461}
}

@misc{li2015coco,
	title        = {Microsoft COCO: Common Objects in Context},
	author       = {Tsung-Yi Lin and Michael Maire and Serge Belongie and Lubomir Bourdev and Ross Girshick and James Hays and Pietro Perona and Deva Ramanan and C. Lawrence Zitnick and Piotr Dollár},
	year         = 2015,
	eprint       = {1405.0312},
	archiveprefix = {arXiv},
	primaryclass = {cs.CV}
}

@misc{zhou2018ade20k,
	title        = {Semantic Understanding of Scenes through the ADE20K Dataset},
	author       = {Bolei Zhou and Hang Zhao and Xavier Puig and Tete Xiao and Sanja Fidler and Adela Barriuso and Antonio Torralba},
	year         = 2018,
	eprint       = {1608.05442},
	archiveprefix = {arXiv},
	primaryclass = {cs.CV}
}

@misc{zhang2016colorization,
	title        = {Colorful Image Colorization},
	author       = {Richard Zhang and Phillip Isola and Alexei A. Efros},
	year         = 2016,
	eprint       = {1603.08511},
	archiveprefix = {arXiv},
	primaryclass = {cs.CV}
}

@misc{hjelm2019dim,
	title        = {Learning deep representations by mutual information estimation and maximization},
	author       = {R Devon Hjelm and Alex Fedorov and Samuel Lavoie-Marchildon and Karan Grewal and Phil Bachman and Adam Trischler and Yoshua Bengio},
	year         = 2019,
	eprint       = {1808.06670},
	archiveprefix = {arXiv},
	primaryclass = {stat.ML}
}

@misc{oord2019cpc,
	title        = {Representation Learning with Contrastive Predictive Coding},
	author       = {Aaron van den Oord and Yazhe Li and Oriol Vinyals},
	year         = 2019,
	eprint       = {1807.03748},
	archiveprefix = {arXiv},
	primaryclass = {cs.LG}
}

@misc{kornblith2019do,
	title        = {Do Better ImageNet Models Transfer Better?},
	author       = {Simon Kornblith and Jonathon Shlens and Quoc V. Le},
	year         = 2019,
	eprint       = {1805.08974},
	archiveprefix = {arXiv},
	primaryclass = {cs.CV}
}

@misc{caron2019deepcluster,
	title        = {Deep Clustering for Unsupervised Learning of Visual Features},
	author       = {Mathilde Caron and Piotr Bojanowski and Armand Joulin and Matthijs Douze},
	year         = 2019,
	eprint       = {1807.05520},
	archiveprefix = {arXiv},
	primaryclass = {cs.CV}
}

@misc{caron2019deepercluster,
	title        = {Unsupervised Pre-Training of Image Features on Non-Curated Data},
	author       = {Mathilde Caron and Piotr Bojanowski and Julien Mairal and Armand Joulin},
	year         = 2019,
	eprint       = {1905.01278},
	archiveprefix = {arXiv},
	primaryclass = {cs.CV}
}

@misc{chen2020simclr,
	title        = {A Simple Framework for Contrastive Learning of Visual Representations},
	author       = {Ting Chen and Simon Kornblith and Mohammad Norouzi and Geoffrey Hinton},
	year         = 2020,
	eprint       = {2002.05709},
	archiveprefix = {arXiv},
	primaryclass = {cs.LG}
}

@misc{grill2020byol,
	title        = {Bootstrap your own latent: A new approach to self-supervised Learning},
	author       = {Jean-Bastien Grill and Florian Strub and Florent Altché and Corentin Tallec and Pierre H. Richemond and Elena Buchatskaya and Carl Doersch and Bernardo Avila Pires and Zhaohan Daniel Guo and Mohammad Gheshlaghi Azar and Bilal Piot and Koray Kavukcuoglu and Rémi Munos and Michal Valko},
	year         = 2020,
	eprint       = {2006.07733},
	archiveprefix = {arXiv},
	primaryclass = {cs.LG}
}

@misc{chen2020mocov2,
	title        = {Improved Baselines with Momentum Contrastive Learning},
	author       = {Xinlei Chen and Haoqi Fan and Ross Girshick and Kaiming He},
	year         = 2020,
	eprint       = {2003.04297},
	archiveprefix = {arXiv},
	primaryclass = {cs.CV}
}

@misc{chen2020simsiam,
	title        = {Exploring Simple Siamese Representation Learning},
	author       = {Xinlei Chen and Kaiming He},
	year         = 2020,
	eprint       = {2011.10566},
	archiveprefix = {arXiv},
	primaryclass = {cs.CV}
}

@misc{he2020moco,
	title        = {Momentum Contrast for Unsupervised Visual Representation Learning},
	author       = {Kaiming He and Haoqi Fan and Yuxin Wu and Saining Xie and Ross Girshick},
	year         = 2020,
	eprint       = {1911.05722},
	archiveprefix = {arXiv},
	primaryclass = {cs.CV}
}

@inproceedings{tian2020infomin,
	title        = {What Makes for Good Views for Contrastive Learning?},
	author       = {Tian, Yonglong and Sun, Chen and Poole, Ben and Krishnan, Dilip and Schmid, Cordelia and Isola, Phillip},
	year         = 2020,
	booktitle    = {Advances in Neural Information Processing Systems},
	publisher    = {Curran Associates, Inc.},
	volume       = 33,
	pages        = {6827--6839},
	url          = {https://proceedings.neurips.cc/paper_files/paper/2020/file/4c2e5eaae9152079b9e95845750bb9ab-Paper.pdf},
	editor       = {H. Larochelle and M. Ranzato and R. Hadsell and M.F. Balcan and H. Lin}
}

@inproceedings{caron2020swav,
	title        = {Unsupervised Learning of Visual Features by Contrasting Cluster Assignments},
	author       = {Caron, Mathilde and Misra, Ishan and Mairal, Julien and Goyal, Priya and Bojanowski, Piotr and Joulin, Armand},
	year         = 2020,
	booktitle    = {Advances in Neural Information Processing Systems},
	publisher    = {Curran Associates, Inc.},
	volume       = 33,
	pages        = {9912--9924},
	url          = {https://proceedings.neurips.cc/paper_files/paper/2020/file/70feb62b69f16e0238f741fab228fec2-Paper.pdf},
	editor       = {H. Larochelle and M. Ranzato and R. Hadsell and M.F. Balcan and H. Lin},
	eprint       = {2006.09882},
	archiveprefix = {arXiv},
	primaryclass = {cs.CV}
}

@misc{kalantidis2020mochi,
	title        = {Hard Negative Mixing for Contrastive Learning},
	author       = {Yannis Kalantidis and Mert Bulent Sariyildiz and Noe Pion and Philippe Weinzaepfel and Diane Larlus},
	year         = 2020,
	eprint       = {2010.01028},
	archiveprefix = {arXiv},
	primaryclass = {cs.CV}
}

@misc{khosla2021supcon,
	title        = {Supervised Contrastive Learning},
	author       = {Prannay Khosla and Piotr Teterwak and Chen Wang and Aaron Sarna and Yonglong Tian and Phillip Isola and Aaron Maschinot and Ce Liu and Dilip Krishnan},
	year         = 2021,
	eprint       = {2004.11362},
	archiveprefix = {arXiv},
	primaryclass = {cs.LG}
}

@misc{chen2021mocov3,
	title        = {An Empirical Study of Training Self-Supervised Vision Transformers},
	author       = {Xinlei Chen and Saining Xie and Kaiming He},
	year         = 2021,
	eprint       = {2104.02057},
	archiveprefix = {arXiv},
	primaryclass = {cs.CV}
}

@misc{zbontar2021barlowtwins,
	title        = {Barlow Twins: Self-Supervised Learning via Redundancy Reduction},
	author       = {Jure Zbontar and Li Jing and Ishan Misra and Yann LeCun and Stéphane Deny},
	year         = 2021,
	eprint       = {2103.03230},
	archiveprefix = {arXiv},
	primaryclass = {cs.CV}
}

@misc{ermolov2021wmse,
	title        = {Whitening for Self-Supervised Representation Learning},
	author       = {Aleksandr Ermolov and Aliaksandr Siarohin and Enver Sangineto and Nicu Sebe},
	year         = 2021,
	eprint       = {2007.06346},
	archiveprefix = {arXiv},
	primaryclass = {cs.LG}
}

@misc{goyal2021seer,
	title        = {Self-supervised Pretraining of Visual Features in the Wild},
	author       = {Priya Goyal and Mathilde Caron and Benjamin Lefaudeux and Min Xu and Pengchao Wang and Vivek Pai and Mannat Singh and Vitaliy Liptchinsky and Ishan Misra and Armand Joulin and Piotr Bojanowski},
	year         = 2021,
	eprint       = {2103.01988},
	archiveprefix = {arXiv},
	primaryclass = {cs.CV}
}

@misc{dwibedi2021nnclr,
	title        = {With a Little Help from My Friends: Nearest-Neighbor Contrastive Learning of Visual Representations},
	author       = {Debidatta Dwibedi and Yusuf Aytar and Jonathan Tompson and Pierre Sermanet and Andrew Zisserman},
	year         = 2021,
	eprint       = {2104.14548},
	archiveprefix = {arXiv},
	primaryclass = {cs.CV}
}

@misc{caron2021dino,
	title        = {Emerging Properties in Self-Supervised Vision Transformers},
	author       = {Mathilde Caron and Hugo Touvron and Ishan Misra and Hervé Jégou and Julien Mairal and Piotr Bojanowski and Armand Joulin},
	year         = 2021,
	eprint       = {2104.14294},
	archiveprefix = {arXiv},
	primaryclass = {cs.CV}
}

@misc{xie2021moby,
	title        = {Self-Supervised Learning with Swin Transformers},
	author       = {Zhenda Xie and Yutong Lin and Zhuliang Yao and Zheng Zhang and Qi Dai and Yue Cao and Han Hu},
	year         = 2021,
	eprint       = {2105.04553},
	archiveprefix = {arXiv},
	primaryclass = {cs.CV}
}

@misc{bardes2022vicreg,
	title        = {VICReg: Variance-Invariance-Covariance Regularization for Self-Supervised Learning},
	author       = {Adrien Bardes and Jean Ponce and Yann LeCun},
	year         = 2022,
	eprint       = {2105.04906},
	archiveprefix = {arXiv},
	primaryclass = {cs.CV}
}

@misc{yeh2022dcl,
	title        = {Decoupled Contrastive Learning},
	author       = {Chun-Hsiao Yeh and Cheng-Yao Hong and Yen-Chi Hsu and Tyng-Luh Liu and Yubei Chen and Yann LeCun},
	year         = 2022,
	eprint       = {2110.06848},
	archiveprefix = {arXiv},
	primaryclass = {cs.LG}
}

@misc{bao2022beit,
	title        = {BEiT: BERT Pre-Training of Image Transformers},
	author       = {Hangbo Bao and Li Dong and Songhao Piao and Furu Wei},
	year         = 2022,
	eprint       = {2106.08254},
	archiveprefix = {arXiv},
	primaryclass = {cs.CV}
}

@misc{xie2022simmim,
	title        = {SimMIM: A Simple Framework for Masked Image Modeling},
	author       = {Zhenda Xie and Zheng Zhang and Yue Cao and Yutong Lin and Jianmin Bao and Zhuliang Yao and Qi Dai and Han Hu},
	year         = 2022,
	eprint       = {2111.09886},
	archiveprefix = {arXiv},
	primaryclass = {cs.CV}
}

@misc{tomasev2022relicv2,
	title        = {Pushing the limits of self-supervised ResNets: Can we outperform supervised learning without labels on ImageNet?},
	author       = {Nenad Tomasev and Ioana Bica and Brian McWilliams and Lars Buesing and Razvan Pascanu and Charles Blundell and Jovana Mitrovic},
	year         = 2022,
	eprint       = {2201.05119},
	archiveprefix = {arXiv},
	primaryclass = {cs.CV}
}

@misc{pang2022smog,
	title        = {Unsupervised Visual Representation Learning by Synchronous Momentum Grouping},
	author       = {Bo Pang and Yifan Zhang and Yaoyi Li and Jia Cai and Cewu Lu},
	year         = 2022,
	eprint       = {2207.06167},
	archiveprefix = {arXiv},
	primaryclass = {cs.CV}
}

@misc{peng2022beitv2,
	title        = {BEiT v2: Masked Image Modeling with Vector-Quantized Visual Tokenizers},
	author       = {Zhiliang Peng and Li Dong and Hangbo Bao and Qixiang Ye and Furu Wei},
	year         = 2022,
	eprint       = {2208.06366},
	archiveprefix = {arXiv},
	primaryclass = {cs.CV}
}

@misc{chen2023cae,
	title        = {Context Autoencoder for Self-Supervised Representation Learning},
	author       = {Xiaokang Chen and Mingyu Ding and Xiaodi Wang and Ying Xin and Shentong Mo and Yunhao Wang and Shumin Han and Ping Luo and Gang Zeng and Jingdong Wang},
	year         = 2023,
	eprint       = {2202.03026},
	archiveprefix = {arXiv},
	primaryclass = {cs.CV}
}

@misc{oquab2023dinov2,
	title        = {DINOv2: Learning Robust Visual Features without Supervision},
	author       = {Oquab, Maxime and Darcet, Timothée and Moutakanni, Theo and Vo, Huy V. and Szafraniec, Marc and Khalidov, Vasil and Fernandez, Pierre and Haziza, Daniel and Massa, Francisco and El-Nouby, Alaaeldin and Howes, Russell and Huang, Po-Yao and Xu, Hu and Sharma, Vasu and Li, Shang-Wen and Galuba, Wojciech and Rabbat, Mike and Assran, Mido and Ballas, Nicolas and Synnaeve, Gabriel and Misra, Ishan and Jegou, Herve and Mairal, Julien and Labatut, Patrick and Joulin, Armand and Bojanowski, Piotr},
	year         = 2023,
	journal      = {arXiv:2304.07193}
}

@misc{yerxa2023mmcr,
	title        = {Learning Efficient Coding of Natural Images with Maximum Manifold Capacity Representations},
	author       = {Thomas Yerxa and Yilun Kuang and Eero Simoncelli and SueYeon Chung},
	year         = 2023,
	eprint       = {2303.03307},
	archiveprefix = {arXiv},
	primaryclass = {cs.CV}
}

@misc{he2021mae,
	title        = {Masked Autoencoders Are Scalable Vision Learners},
	author       = {Kaiming He and Xinlei Chen and Saining Xie and Yanghao Li and Piotr Dollár and Ross Girshick},
	year         = 2021,
	url          = {https://arxiv.org/abs/2111.06377},
	eprint       = {2111.06377},
	archiveprefix = {arXiv},
	primaryclass = {cs.CV}
}

@misc{giakoumoglou2025synco,
	title        = {SynCo: Synthetic Hard Negatives for Contrastive Visual Representation Learning},
	author       = {Nikolaos Giakoumoglou and Tania Stathaki},
	year         = 2025,
	url          = {https://arxiv.org/abs/2410.02401},
	eprint       = {2410.02401},
	archiveprefix = {arXiv},
	primaryclass = {cs.CV}
}

@misc{dosovitskiy2021vit,
	title        = {An Image is Worth 16x16 Words: Transformers for Image Recognition at Scale},
	author       = {Alexey Dosovitskiy and Lucas Beyer and Alexander Kolesnikov and Dirk Weissenborn and Xiaohua Zhai and Thomas Unterthiner and Mostafa Dehghani and Matthias Minderer and Georg Heigold and Sylvain Gelly and Jakob Uszkoreit and Neil Houlsby},
	year         = 2021,
	eprint       = {2010.11929},
	archiveprefix = {arXiv},
	primaryclass = {cs.CV}
}

@misc{loshchilov2019adamw,
	title        = {Decoupled Weight Decay Regularization},
	author       = {Ilya Loshchilov and Frank Hutter},
	year         = 2019,
	eprint       = {1711.05101},
	archiveprefix = {arXiv},
	primaryclass = {cs.LG}
}

@misc{gidaris2018rotnet,
	title        = {Unsupervised Representation Learning by Predicting Image Rotations},
	author       = {Spyros Gidaris and Praveer Singh and Nikos Komodakis},
	year         = 2018,
	eprint       = {1803.07728},
	archiveprefix = {arXiv},
	primaryclass = {cs.CV}
}

@misc{noroozi2017unsupervised,
	title        = {Unsupervised Learning of Visual Representations by Solving Jigsaw Puzzles},
	author       = {Noroozi,  Mehdi and Favaro,  Paolo},
	year         = 2016,
	publisher    = {arXiv},
	doi          = {10.48550/ARXIV.1603.09246},
	url          = {https://arxiv.org/abs/1603.09246},
	copyright    = {arXiv.org perpetual,  non-exclusive license},
	eprint       = {1603.09246},
	archiveprefix = {arXiv},
	primaryclass = {cs.CV}
}

@misc{zhou2022ibot,
	title        = {iBOT: Image BERT Pre-Training with Online Tokenizer},
	author       = {Jinghao Zhou and Chen Wei and Huiyu Wang and Wei Shen and Cihang Xie and Alan Yuille and Tao Kong},
	year         = 2022,
	eprint       = {2111.07832},
	archiveprefix = {arXiv},
	primaryclass = {cs.CV}
}

@misc{kolesnikov2019revisiting,
	title        = {Revisiting Self-Supervised Visual Representation Learning},
	author       = {Alexander Kolesnikov and Xiaohua Zhai and Lucas Beyer},
	year         = 2019,
	eprint       = {1901.09005},
	archiveprefix = {arXiv},
	primaryclass = {cs.CV}
}

@misc{liu2021swin,
	title        = {Swin Transformer: Hierarchical Vision Transformer using Shifted Windows},
	author       = {Ze Liu and Yutong Lin and Yue Cao and Han Hu and Yixuan Wei and Zheng Zhang and Stephen Lin and Baining Guo},
	year         = 2021,
	eprint       = {2103.14030},
	archiveprefix = {arXiv},
	primaryclass = {cs.CV}
}

@misc{balestriero2023cookbook,
	title        = {A Cookbook of Self-Supervised Learning},
	author       = {Randall Balestriero and Mark Ibrahim and Vlad Sobal and Ari Morcos and Shashank Shekhar and Tom Goldstein and Florian Bordes and Adrien Bardes and Gregoire Mialon and Yuandong Tian and Avi Schwarzschild and Andrew Gordon Wilson and Jonas Geiping and Quentin Garrido and Pierre Fernandez and Amir Bar and Hamed Pirsiavash and Yann LeCun and Micah Goldblum},
	year         = 2023,
	eprint       = {2304.12210},
	archiveprefix = {arXiv},
	primaryclass = {cs.LG}
}

@misc{lan2020albert,
	title        = {ALBERT: A Lite BERT for Self-supervised Learning of Language Representations},
	author       = {Zhenzhong Lan and Mingda Chen and Sebastian Goodman and Kevin Gimpel and Piyush Sharma and Radu Soricut},
	year         = 2020,
	eprint       = {1909.11942},
	archiveprefix = {arXiv},
	primaryclass = {cs.CL}
}

@misc{devlin2018bert,
	title        = {BERT: Pre-training of Deep Bidirectional Transformers for Language Understanding},
	author       = {Jacob Devlin and Ming-Wei Chang and Kenton Lee and Kristina Toutanova},
	year         = 2018,
	eprint       = {1810.04805},
	archiveprefix = {arXiv},
	primaryclass = {cs.CL}
}

@article{lecun2015deep,
	title        = {Deep learning},
	author       = {LeCun,  Yann and Bengio,  Yoshua and Hinton,  Geoffrey},
	year         = 2015,
	month        = may,
	journal      = {Nature},
	publisher    = {Springer Science and Business Media LLC},
	volume       = 521,
	number       = 7553,
	pages        = {436–444},
	doi          = {10.1038/nature14539},
	issn         = {1476-4687},
	url          = {http://dx.doi.org/10.1038/nature14539}
}

@misc{ioffe2015batch,
	title        = {Batch Normalization: Accelerating Deep Network Training by Reducing Internal Covariate Shift},
	author       = {Sergey Ioffe and Christian Szegedy},
	year         = 2015,
	eprint       = {1502.03167},
	archiveprefix = {arXiv},
	primaryclass = {cs.LG}
}

@inproceedings{nair2010relu,
	title        = {Rectified Linear Units Improve Restricted Boltzmann Machines Vinod Nair},
	author       = {Nair, Vinod and Hinton, Geoffrey},
	year         = 2010,
	month        = {06},
	journal      = {Proceedings of ICML},
	volume       = 27,
	pages        = {807--814}
}

@inproceedings{radford2018improving,
	title        = {Improving Language Understanding by Generative Pre-Training},
	author       = {Alec Radford and Karthik Narasimhan},
	year         = 2018,
	url          = {https://api.semanticscholar.org/CorpusID:49313245}
}

@inproceedings{radford2019language,
	title        = {Language Models are Unsupervised Multitask Learners},
	author       = {Alec Radford and Jeff Wu and Rewon Child and David Luan and Dario Amodei and Ilya Sutskever},
	year         = 2019,
	url          = {https://api.semanticscholar.org/CorpusID:160025533}
}

@article{brown2020language,
	title        = {Language models are few-shot learners},
	author       = {Brown, Tom and Mann, Benjamin and Ryder, Nick and Subbiah, Melanie and Kaplan, Jared D and Dhariwal, Prafulla and Neelakantan, Arvind and Shyam, Pranav and Sastry, Girish and Askell, Amanda and others},
	year         = 2020,
	journal      = {Advances in neural information processing systems},
	volume       = 33,
	pages        = {1877--1901}
}

@misc{radford2021clip,
	title        = {Learning Transferable Visual Models From Natural Language Supervision},
	author       = {Alec Radford and Jong Wook Kim and Chris Hallacy and Aditya Ramesh and Gabriel Goh and Sandhini Agarwal and Girish Sastry and Amanda Askell and Pamela Mishkin and Jack Clark and Gretchen Krueger and Ilya Sutskever},
	year         = 2021,
	url          = {https://arxiv.org/abs/2103.00020},
	eprint       = {2103.00020},
	archiveprefix = {arXiv},
	primaryclass = {cs.CV}
}

@misc{touvron2021deit,
	title        = {Training data-efficient image transformers and distillation through attention},
	author       = {Hugo Touvron and Matthieu Cord and Matthijs Douze and Francisco Massa and Alexandre Sablayrolles and Hervé Jégou},
	year         = 2021,
	url          = {https://arxiv.org/abs/2012.12877},
	eprint       = {2012.12877},
	archiveprefix = {arXiv},
	primaryclass = {cs.CV}
}

@inproceedings{vaswani2017attention,
	title        = {Attention is all you need},
	author       = {Vaswani, Ashish and Shazeer, Noam and Parmar, Niki and Uszkoreit, Jakob and Jones, Llion and Gomez, Aidan N and Kaiser, {\L}ukasz and Polosukhin, Illia},
	year         = 2017,
	booktitle    = {Advances in Neural Information Processing Systems},
	pages        = {5998--6008}
}

@misc{huang2016deep,
	title        = {Deep Networks with Stochastic Depth},
	author       = {Gao Huang and Yu Sun and Zhuang Liu and Daniel Sedra and Kilian Weinberger},
	year         = 2016,
	url          = {https://arxiv.org/abs/1603.09382},
	eprint       = {1603.09382},
	archiveprefix = {arXiv},
	primaryclass = {cs.LG}
}

@inproceedings{robinson2021contrastive,
	title        = {Contrastive learning with hard negative samples},
	author       = {Robinson, Joshua and Chuang, Ching-Yao and Sra, Suvrit and Jegelka, Stefanie},
	year         = 2021,
	booktitle    = {International Conference on Learning Representations}
}

@misc{balestriero2025lejepa,
	title        = {LeJEPA: Provable and Scalable Self-Supervised Learning Without the Heuristics},
	author       = {Randall Balestriero and Yann LeCun},
	year         = 2025,
	url          = {https://arxiv.org/abs/2511.08544},
	eprint       = {2511.08544},
	archiveprefix = {arXiv},
	primaryclass = {cs.LG}
}

@misc{kuang2026lpjepa,
	title        = {Rectified LpJEPA: Joint-Embedding Predictive Architectures with Sparse and Maximum-Entropy Representations},
	author       = {Yilun Kuang and Yash Dagade and Tim G. J. Rudner and Randall Balestriero and Yann LeCun},
	year         = 2026,
	url          = {https://arxiv.org/abs/2602.01456},
	eprint       = {2602.01456},
	archiveprefix = {arXiv},
	primaryclass = {cs.LG}
}
}

\end{document}